\documentclass[11pt]{article}

\usepackage[american]{babel}
\usepackage[margin=1in]{geometry}
\usepackage[T1]{fontenc}
\usepackage[utf8]{inputenc}
\usepackage{lmodern}
\usepackage{microtype}

\usepackage{natbib} 
\usepackage{mathtools}
\usepackage{booktabs}
\usepackage{tikz}

\usepackage{amsmath}
\usepackage{amssymb}
\usepackage{amsthm}
\usepackage{bbm}
\usepackage{dsfont}
\usepackage{hyperref}
\usepackage[capitalize,noabbrev]{cleveref}
\usepackage{subfigure} 

\theoremstyle{plain}
\newtheorem{theorem}{Theorem}[section]

\newtheorem{lemma}[theorem]{Lemma}
\newtheorem{corollary}[theorem]{Corollary}
\theoremstyle{definition}
\newtheorem{definition}[theorem]{Definition}
\newtheorem{assumption}[theorem]{Assumption}
\theoremstyle{remark}
\newtheorem{remark}[theorem]{Remark}

\def\cA{{\mathcal A}}

\def\cB{{\mathcal B}}
\def\cP{{\mathcal P}}
\newcommand{\bw}{\boldsymbol{w}}

\newcommand{\bX}{\boldsymbol{X}}
\newcommand{\bZ}{\boldsymbol{Z}}
\newcommand{\bx}{\boldsymbol{x}}
\newcommand{\bz}{\boldsymbol{z}}
\newcommand{\bB}{\boldsymbol{B}}
\newcommand{\bR}{\boldsymbol{R}}
\newcommand{\bgamma}{\boldsymbol{\gamma}}
\newcommand{\bbeta}{\boldsymbol{\beta}}
\newcommand{\bphi}{\boldsymbol{\phi}}
\newcommand{\mP}{\mathbb{P}}
\newcommand{\mE}{\mathbb{E}}
\newcommand{\mR}{\mathbb{R}}
\newcommand{\cN}{\mathcal{N}}
\newcommand{\cX}{\mathcal{X}}
\newcommand{\cF}{\mathcal{F}}
\newcommand{\cQ}{\mathcal{Q}}
\newcommand{\cU}{\mathcal{U}}
\newcommand{\cG}{\mathcal{G}}

\newcommand{\Rad}{\operatorname{Rad}}

\title{Targeted Regularization for Causal Effect Estimation\\ with Exponential Dispersion Family Outcomes}

\author{
Jiahong Li\textsuperscript{1}\thanks{Jiahong Li and Zeqin Yang contributed equally.}\thanks{Corresponding author: Jiahong Li \texttt{caleblijiahong@didiglobal.com}.}\thanks{The authors thank Jiayi Dan for providing an earlier version of the proofs in this paper.},
Zeqin Yang\textsuperscript{1}\footnotemark[1],
Jixing Xu\textsuperscript{1},
Enzheng Hua\textsuperscript{1},
Zhichao Zou\textsuperscript{1},
Peng Zhen\textsuperscript{1},
Jiecheng Guo\textsuperscript{1}\\
\textsuperscript{1}Didi Chuxing, Beijing, China
}
\date{}
\begin{document}

\maketitle

\begin{abstract}
    Neural Networks (NNs) for causal effect estimation have shown strong empirical performance, yet endowing them with desirable semiparametric properties---doubly robustness and fast convergence rates---remains challenging.
    A common approach to address this is targeted regularization, which modifies the objective function of NNs.
    However, existing work on neural causal effect estimation is largely limited to continuous outcomes, restricting its applicability to settings involving binary, count, or other skewed outcomes commonly encountered in practice. We propose a unified targeted regularization framework for the Exponential Dispersion Family (EDF) to address this limitation.
    Specifically, we first derive the von Mises expansion of the average dose function of canonical functions (ADCF) for discrete treatments and of the sieve-projected ADCF for continuous treatments.
    Second, we use this expansion to construct a unified targeted regularization, that corrects first-order bias at the distributional level. We integrate this objective into a NN architecture that jointly estimates the outcome model, propensity score model, and fluctuation parameter end-to-end.
    Experimental results demonstrate the effectiveness of our method. 
\end{abstract}

\section{Introduction}

Recently, estimating causal effects from observational data has gained significant attention in various fields \citep{glass2013causal, li2016matching}, which faces the primary challenge of efficient learning.
Meanwhile, Neural Networks (NNs) have already shown strong potential in causal effect estimation, and established several influential paradigms \citep{johansson2018learning, wang2022generalization}.
However, it still remains appealing to design a NN-based estimator for causal effect estimation while achieving desirable properties, such as doubly robustness and fast convergence rates.

Based semiparametric theory, two basic tools have been developed to achieve causal effect estimation with great theoretical properties: the Doubly Robust (DR) estimator \citep{KennedyDR2023} and the Targeted Maximum Likelihood Estimation (TMLE) \citep{van2011targeted}.
The DR estimator corrects bias by estimating and subtracting it from the initial plug-in estimator.
TMLE, alternatively, eliminates bias at the distribution level by constructing a fluctuated estimator that zeroes out the bias term.
Building on the foundation of TMLE, \citep{shi2019adapting} makes a groundbreaking contribution by introducing targeted regularization.
It seamlessly integrating the TMLE theory into the design of NNs, resulting in a NN-based estimator with non-parametrically optimal asymptotic properties for binary treatment setting.
Subsequently, \citep{nie2021vcnet} advanced this framework for continuous treatment by extending the functional targeted regularization.
This approach ensures theoretical soundness in NN estimators through targeted regularization.
\citep{nie2021vcnet} extended this concept to functional targeted regularization, developing a doubly robust and consistent estimator for the whole ADRF curve to handle continuous treatment scenarios.

However, while the above NNs-based estimators \citep{shi2019adapting, nie2021vcnet} have made significant progress, they primarily focus on continuous outcomes, implicitly making Gaussian distribution assumption.
Therefore, these estimators fail to address the binary or count outcomes, which are very common in real-world applications.
For example, in social media advertising, platforms like Instagram show targeted ads to users (treatment) and track whether those users ultimately purchase the advertised product (binary outcome).
Although \citep{gao2022estimatingheterogeneoustreatmenteffects} introduce DINA (Difference In Natural pArameters) to quantify the treatment effect of exponential family outcomes, their extended R-learner framework is limited to the partially linear assumption.

To address these limitations, we propose an end-to-end NN-based estimator for Exponential Dispersion Families (EDF).
Although canonical (natural) parameters have been used to quantify causal effects for exponential family outcomes in \citep{gao2022estimatingheterogeneoustreatmenteffects}, how to design targeted regularization for NNs to correct bias is still remains unexplored.
Following this direction, we must first derive and understand what the bias term is before we construct the debiased and doubly robust estimator.
However, to the best of our knowledge, the existing literature has not addressed bias correction through targeted regularization for Average Dose function of Canonical Functions (ADCF).
To achieve this, we first introduce the von Mises expansion of ADCF to identify the first-order bias term for plug-in estimator, which inspires us how to targeted regularization for exponential dispersion family outcomes and analyze its asymptotic properties.
Leveraging the above theoretical findings, we generalize  using the doubly robust estimator.


Our specific contributions are as follows:
\begin{enumerate}
    \item \textbf{Semiparametric analysis of the ADCF.}
    We derive the von Mises expansion of the ADCF for exponential dispersion family outcomes, characterizing the first-order bias of the plug-in estimator.
    For continuous treatments, where the pointwise ADCF is not pathwise differentiable, we introduce a sieve-projected ADCF that restores pathwise differentiability and admits a well-defined influence function (Lemmas~\ref{vonmiseexpansion} and~\ref{lem:sieve-VonMises}).

    \item \textbf{Unified targeted regularization.}
    Leveraging the bias structure revealed by the von Mises expansion, we construct a unified targeted regularization objective for EDF.
    The resulting NN-based estimator jointly learns the outcome model, propensity score model, and fluctuation parameter end-to-end, correcting bias at the distributional level while remaining within the parameter space.
    We prove a doubly robust convergence rate of $O_p(n^{-1/3}\sqrt{\log n} + r_1(n)\,r_2(n) + r_2(n)^2)$ in $L^2$ norm (Theorem~\ref{targetedreg}), extending the guarantees of \citet{nie2021vcnet} from Gaussian to the full EDF.

    \item \textbf{Empirical validation.}
    We validate the proposed estimator on synthetic and semi-synthetic datasets with Bernoulli and Poisson outcomes under binary and continuous treatments,
    achieving state-of-the-art results.
\end{enumerate}

\section{Problem Statement and Notations}

Suppose we observe a sample $(\bZ_1, \ldots, \bZ_n)$ of independent and identically distributed observations from some distribution $\mP$, where $\bZ_i = (\bX_i, A_i, Y_i)$ comprises the vector of covariates $\bX_i \in \mathcal{X} \subset \mR^d$, the treatment of interest $A_i \in \mathcal{A} \subset \mR$, and the observed outcome $Y_i \in \mathcal{Y} \subset \mathbb{R}$.
Throughout the paper we assume the treatment $A$ is binary with $\mathcal{A} = \{0, 1\}$ or continuous with $\mathcal{A} = [0,1]$.
Additionally, we assume that $Y$ is sampled from a single-parameter Exponetial Dispersion Family (EDF), which can be view as a single-parameter exponential family with nuisance parameters \citep{Wuthrich2022} as follows:
\begin{small}
\begin{equation}
    Y \sim f(y; \theta, \varphi) = \exp \left\{\frac{y \theta - \kappa(\theta)}{\varphi} + \xi(y; \varphi) \right\}.
\end{equation}
\end{small}
Here, $\kappa: \Theta \rightarrow \mR $ is the cumulant function, $\theta \in \Theta$ is the canonical (natural) parameter modeled as $\theta = \theta(\bx, a)$, which could be further expressed as $\theta(\bx, a) = h(\mu(\bx,a))$, where $h(\cdot)$ is the link function in exponential family and $\mu(\bx,a)=\mathbb{E}[Y|\bx, a]$ represents the conditional mean of $Y$ given $\bx$ and $a$.
Additionally, $\varphi > 0$ is the dispersion parameter, and $\xi(\cdot;\cdot)$ is the normalization, which does not depend on the canonical parameter $\theta$.

In this paper, we want to estimate the Average Dose Canonical Function(ADCF) of EDF, which quantify the treatment effect on the canonical (natural) parameter level.
As highlighted by \citep{gao2022estimatingheterogeneoustreatmenteffects}, using the canonical parameter $\theta$ to represent ADCF is advantageous because it aligns with common practices of comparison on the canonical parameter scale, simplifies modeling the influence of covariates, and avoids uninformative heterogeneity often seen in conditional means.
Therefore, we consider the ADCF as the causal estimand, which is given by
\begin{equation}
\psi_a :=\theta[do(A=a)] = h (\mE[Y|do(A=a)]),
\end{equation} 
where $\mathbb{E}[Y|do(A=a)]$ is the expected potential outcome that would have been observed under treatment level $a$.
In addition, suppose the (generalized) propensity score $\pi(a \mid \bx)$ denotes the conditional density of $A$ given $\bX$.

As is typical in causal inference, we make the strong ignorabililty assumptions:
(i) there exists some constant $c > 0$ such that $\pi(a \mid \bx) \geq c$ for all $\boldsymbol{x} \in \mathcal{X}$ and $a \in \mathcal{A}$;
(ii) The measured covariate $\bX$ blocks all backdoor paths bewteen the treatment $A$ and outcomes $Y$.
Under these assumptions, the causal estimand $\psi_a$ is identified from observational data as a statistical estimand that can be viewed as a functional mapping from the true probability measure $\mP$ to $\mR$:
\begin{small}
    \begin{equation}\label{tarpar}
    \psi_a(\mP) = \mathbb{E}[\theta(\bX, A=a)]= \mathbb{E}[h(\mathbb{E}[Y \mid \boldsymbol{X}, A = a])].
    \end{equation}
\end{small}

\paragraph{Notation} We use $\mE$ to denote expectation, $\mP \in \mathcal{P}$ to denote true probability measure and we write $\mP(f) = \int f(\bz) d\mP(z)$, where $\mathcal{P}$ is a set of possible probability distributions.
Similarly, we use $\mP_n$ to denote the empirical measure and we write $\mP_n(f) = \int f(\bz) d\mP_n(z)$.
We denote convergence in distribution by $\overset{d}{\rightarrow}$ and convergence in probability by $\overset{p}{\rightarrow}$.
$X_n = O_{\mP}(r_n)$ means $X_n / r_n$ is bounded in probability and $X_n = o_{\mP}(r_n)$ means $X_n / r_n \overset{p}{\rightarrow} 0$.
We use $\tau$ to denote Rademacher random variables. We denote Rademacher complexity of a function class $\mathcal{F}: \mathcal{X} \to \mathbb{R}$ as $\text{Rad}_n(\mathcal{F}) = \mathbb{E}(\sup_{f\in\mathcal{F}}|\frac{1}{n}\sum_{i=1}^n \tau_i f(X_i)|)$.
Given two functions $f_1, f_2: \mathcal{X} \to \mathbb{R}$, we define $\|f_1-f_2\|_{\infty} = \sup_{x\in\mathcal{X}}|f_1(x)-f_2(x)|$ and $\|f_1-f_2\|_{L^2} = (\int_{x\in\mathcal{X}}(f_1(x)-f_2(x))^2 dx)^{1/2}$.
For a function class $\mathcal{F}$, we define $\|\mathcal{F}\|_{\infty} = \sup_{f\in\mathcal{F}}\|f\|_{\infty}$.
$a_n \asymp b_n$ denotes that both $a_n/b_n$ and $b_n/a_n$ are bounded.

\section{Plug-in Estimator}\label{plugin}

In this section,
we provide a brief overview of the plug-in estimator for $\psi_a$ using NNs. 
According to Eq. (\ref{tarpar}),
a plug-in estimator is given by 
$\psi_a(\hat{\mP})=\frac{1}{n}\sum_{i=1}^n \hat{\theta}_i(\bx_i, a) =\frac{1}{n}\sum_{i=1}^n h(\hat{\mu}(\bx_i, a))$,
where $\hat{\mu}$ is the estimator of $\mu(\bx,a)$ and $h(\cdot)$ is determined by the actual outcome distribution.
Following \cite{shi2019adapting, nie2021vcnet},
we extract the features related to treatments $A$ before downstream estimation of $\mu(\bx,a)$,
which helps reduce noise and is sufficient to estimate $\psi_a$.
Consequently, 
the NN architecture of the plug-in estimator consists of two heads: 
one for estimating the outcomes $\mu(\bx,a)$, 
and the other for estimating the generalized propensity scores $\pi(a \mid \bx)$.

\paragraph{Discrete treatment.}
When the treatment is binary ($\cA = \{0,1\}$), we adopt the multi-head architecture of \citet{shi2019adapting}.
The propensity head estimates $\hat{\pi}(a|\bx)$ via logistic regression on the shared representation, and the outcome head produces separate output heads $\hat{\mu}(\bx, 0)$ and $\hat{\mu}(\bx, 1)$ for each treatment level.
The final activation of the outcome head is chosen to match the exponential family: sigmoid for Bernoulli outcomes and exponential for Poisson outcomes.

\paragraph{Continuous treatment.}
When the treatment is continuous ($\cA = [0,1]$), we adopt the varying-coefficient neural network structure of \citet{nie2021vcnet}.
The key idea is to let the treatment $a$ modulate the parameters of the outcome head via a varying-coefficient model \citep{hastie1993varying, fan1999statistical, chiang2001smoothing}.
Specifically, the weights of the last layer are parameterized as $w(a)=\sum_{l=1}^{L} \alpha_l \phi_l(a)$, where $\alpha_l$ are learnable coefficients and $\phi_l(\cdot)$ are polynomial basis functions.
This ensures that the ADCF produced by $\hat{\mu}(\bx,a)$ is continuous in $a$ whenever the activation function is continuous.
For the generalized propensity score $\pi(a|\bx)$ in the continuous case, we divide $[0, 1]$ equally into $B$ grids and estimate the conditional density at the $(B+1)$ grid points via $\pi_{\text{grid}}(\bx)=\text{softmax}(\bw\bx)\in \mR^{B+1}$, and then obtain the density for any $a$ via linear interpolation: $\hat{\pi}(a|\bx)=\pi_{\text{grid}}^{a_1}(\bx)+B(\pi_{\text{grid}}^{a_2}(\bx)-\pi_{\text{grid}}^{a_1}(\bx))(a-a_1)$, where $a_1=\lfloor Ba\rfloor$ and $a_2=\lceil Ba \rceil$.

\paragraph{Loss function for the plug-in estimator.}
Since the outcomes $Y$ are sampled from an EDF, we use the negative log-likelihood of the EDF as the loss function for the outcome prediction head, rather than the MSE used in prior works \citep{shi2019adapting, nie2021vcnet}.
For the propensity score head, the negative log-likelihood of $\pi(a|\bx)$ is used.
The total loss for the plug-in estimator is
\begin{equation}\label{Loss}
    \mathcal{L}(\mu, \pi) = \frac{1}{n}\sum_{i=1}^n \left[ \ell(y_i, \mu(\bx_i, a_i)) - \log(\pi(a_i|\bx_i)) \right],
\end{equation}
where $\ell(y, \mu) = -[y\,h(\mu) - \kappa(h(\mu))]/\varphi$ is the negative log-likelihood contribution from the EDF (up to terms not depending on $\mu$).
After obtaining $\hat{\mu}(\bx, a)$ by minimizing Eq. (\ref{Loss}),
the plug-in estimator is $\psi_a(\hat{\mP})=\frac{1}{n}\sum_{i=1}^n \hat{\theta}_i(a, \bx_i) =\frac{1}{n}\sum_{i=1}^n h(\hat{\mu}(a, \bx_i))$.
However, 
the above plug-in estimator $\psi_a(\hat{\mP})$ generally fails to achieve $\sqrt{n}$-consistency.
In the next section, we show how to combine $\hat{\mu}$ and $\hat{\pi}$ through targeted regularization to correct this bias at the distributional level.


\section{Methodology}\label{methodology}

In this section, we present a unified methodology for estimating the ADCF of EDF under both discrete and continuous treatments.
We begin in Section~\ref{doublyrobustestimator} by characterizing the first-order bias of the plug-in estimator through a von Mises expansion for discrete treatments.
Section~\ref{sec:sieve} addresses the well-known obstacle that the pointwise ADCF is not pathwise differentiable under continuous treatment, and introduces the sieve-projected ADCF as a smooth surrogate target.
These theoretical insights then guide the design of a unified NN architecture in Section~\ref{plugin} and, subsequently, a targeted regularization objective that corrects bias at the distributional scale in Section~\ref{tr}.
Finally, Section~\ref{example} instantiates the framework for two concrete exponential family members: the Bernoulli and Poisson distributions.

\subsection{Discrete Treatment Estimation}\label{doublyrobustestimator}

We begin by examining why the naive plug-in estimator for the ADCF is biased and how the von Mises expansion reveals the structure of this bias.

\paragraph{Bias of the plug-in estimator.}
Given the statistical estimand in Eq.~(\ref{tarpar}), a natural first step is to replace the unknown distribution $\mP$ with an empirical estimate $\hat{\mP}$, yielding the plug-in estimator $\psi_a(\hat{\mP}) = \mP_n[h(\hat{\mu}(\bX, a))]$.
However, such an estimator typically introduces non-negligible bias.
Specifically, when $\hat{\mu}$ is obtained by minimizing an empirical risk, the plug-in estimator inherits both the approximation error of the function class and the finite-sample bias from fitting the conditional mean; these two sources of error are generally of order $O_{\mP}(n^{-s})$ for some $s < 1/2$ and cannot achieve $\sqrt{n}$-consistency on their own.
To understand the structure of this bias and lay the groundwork for its correction, we derive the von Mises expansion of $\psi_a(\mP)$.

\begin{lemma}[Von Mises expansion for discrete treatment]\label{vonmiseexpansion}
Let $\psi_a(\mP) = \mathbb{E}\left\{h(\mathbb{E}[Y \mid \boldsymbol{X}, A = a]) \right\}$ for a twice continuously differentiable link function $h$.
For another probability measure $\bar{\mP}$, the functional $\psi_a$ admits the von Mises expansion:
\begin{small}
    \begin{equation}
    \label{von}
        \psi_a(\bar{\mP}) - \psi_a(\mP) = \int \phi_a(\bar{\mP})\, d (\bar{\mP}-\mP) + R_2(\bar{\mP},\mP),
    \end{equation}
\end{small}

where $\int \phi_a(\bar{\mP})\, d (\bar{\mP}-\mP)$ is shorthand for $\int \phi_a(\bar{\mP})\, d \bar{\mP} - \int \phi_a(\bar{\mP})\, d \mP$, the \emph{efficient influence function} is
{\fontsize{8.5pt}{10pt}\selectfont
\begin{align}
\phi_a(\mP)
&= \frac{\mathds{1}(A=a)}{\pi(a \mid \bx)}
\left\{ Y - \mu(\bx,a)\right\} h^{\prime} (\mu(\bx,a))
+ h (\mu(\bx, a)) - \psi_a(\mP), \nonumber
\end{align}
}
and the second-order remainder is
\begin{small}
\begin{align}
    R_2(&\bar{\mP},\mP) =
     \frac{1}{2} \int h^{\prime \prime}(\mu^*(\bx,a)) \left[\bar{\mu}(\bx,a)-\mu(\bx,a)\right]^2 d \mP(x)\nonumber \\
    & +\int \left\{\frac{\pi(a|\bx)}{\bar{\pi}(a|\bx)} -1\right\} h^{\prime} \left[\bar{\mu}(\bx,a)\right](\mu(\bx,a) - \bar{\mu}(\bx,a))\, d \mP(x), \nonumber
\end{align}
\end{small}
where $(\mu(\bx,a), \pi(a|\bx))$ and $(\bar{\mu}(\bx,a), \bar{\pi}(a|\bx))$ denote the mean function and propensity score under $\mP$ and $\bar{\mP}$ respectively, and $\mu^*(\bx,a)$ is an intermediate value between $\mu(\bx,a)$ and $\bar{\mu}(\bx, a)$.
\end{lemma}

\paragraph{From bias structure to targeted regularization.}
The von Mises expansion in Lemma~\ref{vonmiseexpansion} reveals two critical features that motivate targeted regularization.
First, the remainder $R_2(\bar{\mP},\mP)$ has a \emph{product structure}: it involves only products of nuisance estimation errors $(\bar{\mu} - \mu)$ and $(\bar{\pi}/\pi - 1)$.
Therefore, $R_2$ is second-order small whenever at least one nuisance function is consistently estimated---this is the doubly robust property.
Second, since $\int \phi_a(\hat{\mP})\,d\hat{\mP} = 0$ (the influence function has mean zero under $\hat{\mP}$; \citealt{kennedy2023semiparametricdoublyrobusttargeted}), setting $\bar{\mP} = \hat{\mP}$ yields
\begin{equation*}
    \psi_a(\hat{\mP}) - \psi_a(\mP) = -\int \phi_a(\hat{\mP})\, d \mP + R_2(\hat{\mP},\mP).
\end{equation*}
By replacing the population distribution $\mP$ with the empirical measure $\mP_n$, the bias-corrected \emph{doubly robust (DR)} estimator is obtained:
\begin{equation}\label{onestep}
{\fontsize{8.5pt}{10pt}\selectfont
\begin{aligned}
\hat{\psi}_a^{\text{dr}} =& \psi_a(\hat{\mP}) + \mP_n(\phi_a(\hat{\mP})) \\
    =& \mP_n \left[h(\hat{\mu}(a,\bx))\right]+\mP_n \left[\frac{\mathds{1}(A=a)}{\hat{\pi}(a \mid \bx)} \left\{ Y - \hat{\mu}(\bx,a)\right\} h^{\prime} (\hat{\mu}(\bx,a))\right].
\end{aligned}
}
\end{equation}
The asymptotic behavior of this estimator follows from the decomposition:
\begin{small}
    \begin{equation}
   \begin{aligned}
    \hat{\psi}_a^{\text{dr}} - \psi_a(\mP) &= (\mP_n - \mP)\left\{ \phi_a(\mP) \right\} \\
    &+ (\mP_n - \mP)\left\{ \phi_a(\hat{\mP}) - \phi_a(\mP) \right\}
    + R_2(\hat{\mP}, \mP),
\end{aligned} 
\end{equation}
\label{decomposition}
\end{small}

where $\phi_a(\mP)$ and $R_2(\hat{\mP}, \mP)$ are given by Lemma~\ref{vonmiseexpansion}.

A direct consequence of the product structure of $R_2$ is that the DR estimator achieves $\sqrt{n}$-consistency under mild convergence conditions on the nuisance estimators.

\begin{corollary}\label{asydistr}
If
(1) $(\mP_n - \mP)\left\{ \phi_a(\hat{\mP}) - \phi_a(\mP) \right\} = o_{\mP}(1/\sqrt{n})$;\;
(2) $\Vert \hat{\pi}(a|\bx) - \pi(a|\bx) \Vert = o_{\mP}(n^{-1/4})$;
and (3) $\Vert \hat{\mu}(a,\bx) - \mu(a,\bx) \Vert = o_{\mP}(n^{-1/4})$ hold,
then we have $\hat{\psi}_a^{\text{dr}} - \psi_a(\mP) = (\mP_n - \mP)\left\{ \phi_a(\mP) \right\} + o_{\mP}(1/\sqrt{n})$, and so $\hat{\psi}_a^{\text{dr}}$ is $\sqrt{n}$-consistent and asymptotically normal.
\end{corollary}

\begin{remark}
The first term of Eq.~(\ref{von}) is a simple average of a fixed function, which converges to a normal distribution by the central limit theorem.
In Corollary \ref{asydistr},
Condition~(1) can be ensured via sample splitting \citep{kennedy2023semiparametricdoublyrobusttargeted}.
Conditions~(2) and~(3) can be satisfied by various nonparametric estimators, including random forests and neural networks \citep{Chernozhukov2018, Farrell2021}.
\end{remark}

While the DR estimator provides a principled debiasing strategy, it directly adds the correction term to the plug-in estimate, which can push the estimator outside the parameter space (e.g., outside $[0,1]$ for probabilities).
This motivates the targeted regularization developed in Section~\ref{tr}, which instead corrects bias at the \emph{distributional level}.

\subsection{Continuous Treatment Estimation via Sieve Projection}\label{sec:sieve}

\paragraph{The non-pathwise differentiability obstacle.}
When the treatment $A$ is continuous with $\cA = [0,1]$, the pointwise ADCF $a \mapsto \psi_a(\mP)$ is an infinite-dimensional curve rather than a finite collection of scalars.
A fundamental difficulty is that the pointwise functional $\psi_a(\mP) = \mE[h(\mu(\bX, a))]$ is \emph{not} pathwise differentiable at each fixed $a$ \citep{KenndySCDE, kennedy2023semiparametricdoublyrobusttargeted}.
Intuitively, the indicator $\mathds{1}(A = a)$ in the influence function of Lemma~\ref{vonmiseexpansion} has probability zero under a continuous treatment, so the variance of the influence function diverges.
Therefore, the direct analog of the DR estimator is not $\sqrt{n}$-consistent for continuous $A$, and the von Mises expansion of Lemma~\ref{vonmiseexpansion} does not directly apply.

\paragraph{Solution: sieve-projected ADCF.}
To recover pathwise differentiability and a well-defined influence function, we project the ADCF curve onto a finite-dimensional sieve space.
This approach replaces the infinite-dimensional, non-pathwise-differentiable target with a smooth, finite-dimensional surrogate that admits standard semiparametric inference.
We formalize the required conditions as follows.

\begin{assumption}[Regularity for continuous treatments]\label{as:sieve}
Let $\cA=[0,1]$ and suppose:
\begin{enumerate}
    \item[\textup{(S1)}] \textbf{(Smoothness of the ADCF.)}
    The function $a \mapsto \psi(a) = \mE[h(\mu(\bX, a))]$ belongs to the H\"older class $\Sigma(\ell, C_{\Sigma})$ on $[0,1]$ for some integer $\ell \geq 2$ and constant $C_{\Sigma} > 0$.
    That is, $\psi$ is $(\ell-1)$-times continuously differentiable and its $(\ell-1)$-th derivative is Lipschitz.

    \item[\textup{(S2)}] \textbf{(Overlap for continuous treatment.)}
    There exist constants $0 < c_\pi \leq C_\pi < \infty$ such that $c_\pi \leq \pi(a\mid\bx) \leq C_\pi$ for all $a \in \cA$, $\bx \in \mathcal{X}$.
    The marginal density $p(a) = \int \pi(a\mid\bx)\,d\mP(\bx)$ satisfies $c_\pi \leq p(a) \leq C_\pi$ for all $a \in \cA$.

    \item[\textup{(S3)}] \textbf{(B-spline sieve space.)}
    $\mathcal{B}_{K_n} = \operatorname{span}\{B_1,\ldots,B_{K_n}\}$ is the space of polynomial splines of fixed degree $m \geq \ell$ with $K_n$ equally spaced interior knots on $[0,1]$ and dimension $K_n \asymp n^{1/(2\ell+1)}$.
    The Gram matrix $G \in \mR^{K_n \times K_n}$ with entries $G_{jk} = \int_{\cA} B_j(a)\,B_k(a)\,da$ is positive definite.

    \item[\textup{(S4)}] \textbf{(Link function regularity.)}
    The link function $h$ is twice continuously differentiable, and there exist constants $M_1, M_2 > 0$ such that $|h'(y)| \leq M_1$ and $|h''(y)| \leq M_2$ for all $y$ in the range of $\mu$.

    \item[\textup{(S5)}] \textbf{(Conditional moment regularity.)}
    The functions $\bx \mapsto \mu(\bx, a)$ and $\bx \mapsto \pi(a\mid\bx)$ are measurable and bounded uniformly in $a$, and $\mE[Y^2 \mid \bX=\bx, A=a] \leq C_Y$ for some $C_Y > 0$ uniformly.
\end{enumerate}
\end{assumption}

\begin{remark}
Condition~(S1) is the standard smoothness requirement that governs B-spline approximation quality: for $\psi \in \Sigma(\ell, C_\Sigma)$ and splines of degree $m \geq \ell$, the $L^\infty$ sieve approximation error is $O(K_n^{-\ell})$ \textup{\cite{schumaker2007spline,deBoor2001}}.
Condition~(S2) strengthens the pointwise overlap to uniform overlap over $\cA$; this is standard in the dose--response literature \textup{\cite{kennedy2017nonparametric,Chernozhukov2018}}.
Condition~(S3) specifies the sieve and makes explicit the relationship $K_n \asymp n^{1/(2\ell+1)}$ that balances approximation error against estimation variance; the special case $K_n \asymp n^{1/6}$ corresponds to $\ell = 2$.
Conditions~(S4)--(S5) collect the boundedness requirements that ensure the influence function has finite variance.
\end{remark}

\begin{definition}[Sieve-projected ADCF]\label{def:sieve-ADCF}
Let $\boldsymbol{B}(a) = (B_1(a),\ldots,B_{K_n}(a))^\top$ denote the B-spline basis vector, and let $G = \int_{\cA} \boldsymbol{B}(a)\,\boldsymbol{B}(a)^\top da$ be the Gram matrix.
Define the \emph{sieve coefficient functional}
\begin{small}
    \begin{equation}\label{eq:gamma}
    \boldsymbol{\gamma}(\mP) \;=\; \bigl(\gamma_1(\mP),\ldots,\gamma_{K_n}(\mP)\bigr)^\top,
    \;\; \gamma_k(\mP) \;=\; \int_{\cA} \psi_a(\mP)\,B_k(a)\,da, \nonumber
\end{equation}
\end{small}

and the \emph{sieve-projected ADCF}
\begin{equation}\label{eq:sieve-ADCF}
    \psi^s_a(\mP) \;=\; \bB(a)^\top\,G^{-1}\,\bgamma(\mP)
    \;=\; \sum_{k=1}^{K_n} \beta_k(\mP)\,B_k(a),
\end{equation}
where $\bbeta(\mP) = G^{-1}\bgamma(\mP)$ is the vector of $L^2(\cA)$-projection coefficients of $\psi(\cdot\,;\mP)$ onto $\mathcal{B}_{K_n}$.
\end{definition}

By construction, $\psi^s_a(\mP)$ agrees with the true ADCF up to the sieve approximation error.
Under (S1) and (S3), standard B-spline theory gives
$
\sup_{a\in\cA}\bigl|\psi^s_a(\mP) - \psi_a(\mP)\bigr|
= O\bigl(K_n^{-\ell}\bigr).
$
Crucially, unlike the pointwise ADCF, each sieve coefficient $\gamma_k(\mP)$ \emph{is} pathwise differentiable.
The following lemma formalizes this and establishes the von Mises expansion at the level of the projected curve.

\begin{lemma}[Von Mises expansion for sieve-projected ADCF]\label{lem:sieve-VonMises}
Under Assumption~\ref{as:sieve}, let $h$ be the link function of the exponential dispersion family.
For each $k = 1,\ldots,K_n$, the sieve coefficient functional $\gamma_k(\mP)$ is pathwise differentiable at $\mP$ with influence function
\begin{align}\label{eq:sieve-IF}
    \phi^s_k(\bZ;\mP)
    \;&=\;
    \frac{B_k(A)}{\pi(A \mid \bx)}
    \bigl\{Y - \mu(\bx, A)\bigr\}\,h'\bigl(\mu(\bx, A)\bigr) \nonumber
    \\&+
    \int_{\cA} h\bigl(\mu(\bx, a)\bigr)\,B_k(a)\,da
    \;-\; \gamma_k(\mP). \nonumber
\end{align}
Moreover, $\phi^s_k \in L^2(\mP)$ and $\mE[\phi^s_k(\bZ;\mP)] = 0$.

Consequently, for any probability measure $\bar{\mP}$, each $\gamma_k$ admits the von Mises expansion
\begin{small}
    \begin{equation}\label{eq:sieve-vonMises}
    \gamma_k(\bar{\mP}) - \gamma_k(\mP)
    \;=\;
    \int \phi^s_k(\bz;\bar{\mP})\,d(\bar{\mP}-\mP)(\bz)
    \;+\; R_{2,k}(\bar{\mP},\mP), 
\end{equation}
\end{small}

where the second-order remainder is
\begin{small}
    \begin{align} \label{eq:sieve-remainder}
    &\quad\;\; R_{2,k}(\bar{\mP},\mP)\nonumber\\
    &=\;
    \frac{1}{2}\int_{\cA}\!\!\int_{\mathcal{X}}
    h''\bigl(\mu^*(\bx,a)\bigr)
    \bigl[\bar{\mu}(\bx,a) - \mu(\bx,a)\bigr]^2\,
    B_k(a)\,d\mP(\bx)\,da \nonumber\\
    &+ \int_{\cA}\!\!\int_{\mathcal{X}}
    \left\{\frac{\pi(a|\bx)}{\bar{\pi}(a|\bx)}-1\right\}
    h'\bigl(\bar{\mu}(\bx,a)\bigr)
    \bigl(\mu(\bx,a) - \bar{\mu}(\bx,a)\bigr)\,
    B_k(a)\,d\mP(\bx)\,da,
    \end{align}
\end{small}

with $\mu^*(\bx,a)$ an intermediate value between $\mu(\bx,a)$ and $\bar{\mu}(\bx,a)$.
At the level of the full projected curve, for each $a \in \cA$,
\begin{equation}\label{eq:curve-vonMises}
    \psi^s_a(\bar{\mP}) - \psi^s_a(\mP)
    \;=\;
    \int \phi^s_a(\bz;\bar{\mP})\,d(\bar{\mP}-\mP)(\bz)
    \;+\; R^s_{2}(a;\bar{\mP},\mP), \nonumber
\end{equation}
where the curve-level influence function and remainder are the linear reconstructions
\begin{align}\label{eq:curve-IF}
    \phi^s_a(\bZ;\mP)
    &\;=\;
    \bB(a)^\top G^{-1}\,\bphi^s(\bZ;\mP) \nonumber,
    \\
    R^s_{2}(a;\bar{\mP},\mP)
    &\;=\;
    \bB(a)^\top G^{-1}\,\bR_2(\bar{\mP},\mP) \nonumber,
\end{align}
with $\bphi^s = (\phi^s_1,\ldots,\phi^s_{K_n})^\top$,
$\bR_2 = (R_{2,1},\ldots,R_{2,K_n})^\top$.
\end{lemma}

\begin{remark}
[Doubly robust structure of the remainder]
The second-order remainder~\eqref{eq:sieve-remainder} inherits the product structure from the pointwise remainder in Lemma~\ref{vonmiseexpansion}.
In particular, $R_{2,k}(\bar{\mP},\mP)$ involves only products of nuisance estimation errors $(\bar{\mu}-\mu)$ and $(\bar{\pi}/\pi-1)$, so it vanishes whenever either nuisance function is estimated consistently.
This implies a doubly robust estimator for each sieve coefficient,
\begin{equation}\label{eq:dr-curve}
{\fontsize{8.5pt}{10pt}\selectfont
\begin{aligned}
\hat{\gamma}_k^{\textup{dr}}
&=\mP_n\!\left[
\frac{B_k(A)}{\hat{\pi}(A\mid\bx)}
\bigl\{Y - \hat{\mu}(\bx,A)\bigr\}\,h'(\hat{\mu}(\bx,A))
\;+\; \int_{\cA} h(\hat{\mu}(\bx,a))\,B_k(a)\,da
\right].
\end{aligned}
}
\end{equation}

whose bias depends on~\eqref{eq:sieve-remainder} and is therefore second-order.
\end{remark}

\subsection{Unified Targeted Regularization}\label{tr}

\paragraph{Motivation of TMLE.}
As discussed in \citet{kennedy2023semiparametricdoublyrobusttargeted}, a key limitation of the DR estimator is that even when $\psi_a(\mP)$ and $\psi_a(\hat{\mP})$ are bounded, the additive correction term $\mP_n(\phi_a(\hat{\mP}))$ may push the estimator outside the parameter space.
TMLE \citep{van2011targeted} is an alternative strategy that corrects bias on the \emph{distributional} scale: it constructs a fluctuated estimate $\hat{\mP}^*$ for which the score equation $\mP_n(\phi_a(\hat{\mP}^*)) \approx 0$ is approximately solved.
Building on TMLE, targeted regularization \citep{shi2019adapting, nie2021vcnet} learns the fluctuated $\hat{\mu}^*$ by incorporating an additional regularization objective directly into the NN.

\paragraph{Fluctuation model.}
Targeted regularization introduces a scalar perturbation function $\epsilon: \cA \to \mR$ that defines the fluctuated canonical parameter:
\begin{equation}\label{eq:fluctuated}
    \hat{\theta}^*(\bx, a) \;=\; h(\hat{\mu}(\bx, a)) + \frac{\epsilon(a)}{\hat{\pi}(a|\bx)}\,h'(\hat{\mu}(\bx, a)).
\end{equation}
The goal is to learn $\epsilon$ jointly with $(\hat{\mu}, \hat{\pi})$ such that, at convergence, the score condition
\begin{small}
\begin{align}
\label{tmle}
    \mP_n \left[\frac{\mathds{1}(A=a)}{\hat{\pi}(a \mid \bx)} \left\{ Y - \hat{\mu}^*(\bx,a)\right\} h^{\prime} (\hat{\mu}^*(\bx,a))\right] \approx 0
\end{align}
\end{small}
is approximately satisfied.
When~(\ref{tmle}) holds, the targeted regularization estimator
\begin{small}
    \begin{align}
\label{new}
    \hat{\psi}_a^{\text{tr}} &= \frac{1}{n} \sum_{i=1}^{n} h(\hat{\mu}^*(\bx_i, a))\nonumber\\
    &= \frac{1}{n} \sum_{i=1}^{n} \left(h(\hat{\mu}(\bx_i, a)) + \frac{\hat{\epsilon}_n(a)}{\hat{\pi}(a \mid \bx_i)}h^\prime(\hat{\mu}(\bx_i, a)) \right)
\end{align}
\end{small}
behaves like the DR estimator~(\ref{onestep}) asymptotically, while remaining within the parameter space.

\paragraph{Unified loss function for EDF.}
To enforce the score condition~(\ref{tmle}), we construct a targeted regularization term $\mathcal{R}(\mu, \pi, \epsilon)$ whose derivative with respect to $\epsilon$ equals the score.
The key insight is that for \emph{any} member of EDF, the negative log-likelihood of the fluctuated canonical parameter yields exactly this property.
We define the full loss as
\begin{equation}\label{FTR}
    \mathcal{L}_{\text{TR}}(\mu, \pi, \epsilon) = \mathcal{L}(\mu, \pi) + \beta\, \mathcal{R}(\mu, \pi, \epsilon),
\end{equation}
where $\mathcal{L}(\mu, \pi)$ is defined in Eq.~(\ref{Loss}), $\beta > 0$ is a hyperparameter, and the targeted regularization term is
\begin{small}
\begin{align}
\label{tr_term}
    \mathcal{R}(\mu, \pi, \epsilon) =& \frac{1}{n}\sum_{i=1}^n \left\{-y_i\left[h(\mu(\bx_i, a_i))\right.\right.
    +\left.\left.\frac{\epsilon(a_i)}{\pi(a_i|\bx_i)} h^{\prime}( \mu (\bx_i, a_i))\right]\right.
\nonumber
    \\+&\left.\kappa\left(h(\mu(\bx_i, a_i))+\frac{\epsilon(a_i)}{\pi(a_i|\bx_i)} h^{\prime}( \mu (\bx_i, a_i))\right)\right\},
\end{align}
\end{small}
where $\kappa$ is the cumulant function of the EDF.
By construction, $\mathcal{R}$ is the negative log-likelihood of the fluctuated model $f(y; \hat{\theta}^*, \varphi)$ (up to terms independent of $\epsilon$), so its stationarity in $\epsilon$ is equivalent to~(\ref{tmle}).
Importantly, the form~(\ref{tr_term}) is \emph{unified} across the entire exponential dispersion family: the only distribution-specific ingredient is the cumulant function $\kappa(\cdot)$ and the link function $h(\cdot)$.

For continuous treatments, optimizing $\epsilon(a)$ over all functions $\cA \to \mR$ is infeasible.
Following \citet{nie2021vcnet}, we parameterize $\epsilon$ via the same B-spline sieve $\{B_k\}_{k=1}^{K_n}$ used in Definition~\ref{def:sieve-ADCF}: $\epsilon(a) = \sum_{k=1}^{K_n} \alpha_k B_k(a)$, so that the perturbation is consistent with the sieve projection.

\paragraph{Convergence rate.}
We now state the main theoretical result, which establishes the convergence rate of the targeted regularization estimator under the following assumptions.

\begin{assumption}[Regularity conditions]\label{assumption}
    We consider the following assumptions:
    \begin{enumerate}
        \item[(i)] There exists constant $c > 0$ such that for any $a \in \mathcal{A}$, $\bx \in \mathcal{X}$, and $\hat{\pi} \in \mathcal{Q}$, we have $1/c \leq \hat{\pi}(a\mid \bx) \leq c$, $1/c \leq \pi(a\mid \bx) \leq c$, $\|\mathcal{Q}\|_{\infty} \leq c$ and $\|\mu\|_{\infty} \leq c$.
        \item[(ii)] $\pi, \mu, \hat{\pi}$ and $\hat{\mu}$ have bounded second derivatives for any $\hat{\pi} \in \mathcal{Q}$ and $\hat{\mu} \in \mathcal{U}$.
        \item[(iii)] Either $\hat{\pi} = \pi$ or $\hat{\mu} = \mu$. And $\operatorname{Rad}_n(\mathcal{G}), \operatorname{Rad}_n(\mathcal{Q}), \operatorname{Rad}_n(\mathcal{U}) = O_{\mP}(n^{-1/2})$.
        \item[(iv)] $\mathcal{B}_{K_n}$ equals the closed linear span of B-splines with equally spaced knots, fixed degree, and dimension $K_n \asymp n^{1/6}$.
    \end{enumerate}
\end{assumption}

\begin{theorem} \label{targetedreg}
    Under Assumption~\ref{assumption}, let 
    \begin{small}
        \begin{equation*}
    \hat{\psi}_a^{\text{tr}} = \frac{1}{n} \sum_{i=1}^{n} \left(h(\hat{\mu}(\bx_i, a)) + \frac{\hat{\epsilon}_n(a)}{\hat{\pi}(a \mid \bx_i)}h^\prime(\hat{\mu}(\bx_i, a)) \right)
    \end{equation*} 
    \end{small}
    be the estimator obtained by minimizing $\mathcal{L}_{\text{TR}}$ in Eq.~(\ref{FTR}).
    Then
    \begin{small}
        \[
    \|\hat{\psi}_a^{\text{tr}} - \psi_a\|_{L^2} = O_p\!\left(n^{-\frac{1}{3}}\sqrt{\log n}+r_1(n)\,r_2(n)+r_2(n)^2\right),
    \]
    \end{small}
    where $\|\hat{\pi} - \pi\|_{\infty} = O_p(r_1(n))$ and $\|\hat{\mu} - \mu\|_{\infty} = O_p(r_2(n))$.
\end{theorem}


\begin{remark}
Theorem~\ref{targetedreg} demonstrates \emph{doubly robust convergence}: when both $\hat{\pi}$ and $\hat{\mu}$ converge to their true values, the estimator $\hat{\psi}_a^{\text{tr}}$ achieves a convergence rate faster than either nuisance estimator alone, since $r_1(n)\,r_2(n)$ can dominate slower individual rates.
The rate $n^{-1/3}\sqrt{\log n}$ in the first term reflects the sieve approximation error $O(K_n^{-\ell}) = O(n^{-1/3})$ (for $\ell=2$ and $K_n \asymp n^{1/6}$), combined with the estimation variance $O(\sqrt{K_n/n}) = O(n^{-5/12})$.
This result extends the convergence guarantees of \citet{nie2021vcnet} from Gaussian distributions to the full EDF.
\end{remark}

\begin{remark}[Connection to Lemma~\ref{lem:sieve-VonMises}]
Theorem~\ref{targetedreg} uses the B-spline sieve for the perturbation function $\epsilon(a)=\sum_{k=1}^{K_n}\alpha_k B_k(a)$.
Lemma~\ref{lem:sieve-VonMises} clarifies the deeper role of this sieve: it is not merely a computational convenience for parameterizing $\epsilon$, but a theoretical necessity that renders the ADCF curve a pathwise differentiable statistical target, enabling the existence of the influence function that underpins the targeted regularization.
\end{remark}

\subsection{Exponential Family Examples}\label{example}

We now instantiate the unified framework for two concrete members of the exponential dispersion family: the Bernoulli and Poisson distributions.
These examples illustrate how the general targeted regularization term~(\ref{tr_term}) specializes through the choice of cumulant function $\kappa$ and link function $h$.

\subsubsection{Bernoulli Distribution}

Suppose the outcome $Y \in \{0, 1\}$ follows a Bernoulli distribution.
The canonical link is the logit function $h(\mu)= \log \left(\frac{\mu}{1-\mu}\right)$, and the cumulant function is $\kappa(\theta)=\log(1+\exp(\theta))$.
We use sigmoid in the last layer of the outcome head, yielding $\hat{\mu}(\bX, a)\in (0, 1)$.
The ADCF specializes to
\begin{small}
    \begin{align*}
    \psi_a(\mP) = \mathbb{E}\left\{ \log \frac{\mathbb{E}[Y \mid \boldsymbol{X}, A = a]}{1-\mathbb{E}[Y \mid \boldsymbol{X}, A = a]} \right\},
\end{align*}
\end{small}
i.e., the average log-odds under intervention $do(A=a)$.
From Lemma~\ref{vonmiseexpansion}, the efficient influence function is
\begin{small}
\begin{align*}
    \phi_a(\mP) =& \frac{\mathds{1}(A=a)(Y - \mu(\bx,a))}{\pi(a \mid \bx)\mu(\bx,a)(1-\mu(\bx,a))} + \log \left(\frac{\mu(\bx,a)}{1-\mu(\bx,a)}\right)   - \psi_a,
\end{align*}
\end{small}
where $h'(\mu) = [\mu(1-\mu)]^{-1}$ for the logit link.

Substituting $\kappa(\theta) = \log(1 + e^\theta)$ and $h'(\mu) = [\mu(1-\mu)]^{-1}$ into the general targeted regularization term~(\ref{tr_term}), we obtain
\begin{small}
    \begin{equation*}
    \begin{aligned}
        \mathcal{R}(&\mu, \pi, \epsilon) = \frac{1}{n}\sum_{i=1}^n\left\{-y_i\left[\log\frac{\mu(\bx_i, a_i)}{1-\mu(\bx_i, a_i)}\right.\right.
\\+&\left.\left.\frac{\epsilon(a_i)}{\pi(a_i|\bx_i)} \frac{1}{\mu(\bx_i, a_i)(1-\mu(\bx_i, a_i))}\right]\right.
    \\+&\left.\kappa\left(\log\frac{\mu(\bx_i, a_i)}{1-\mu(\bx_i, a_i)}+\frac{\epsilon(a_i)}{\pi(a_i|\bx_i)} \frac{1}{\mu(\bx_i, a_i)(1-\mu(\bx_i, a_i))}\right)\right\}.
\end{aligned}
\end{equation*}
\end{small}

The argument to $\kappa$ is the fluctuated log-odds, which the neural network can compute end-to-end.
Note that the denominator $\mu(1-\mu)$ in the correction term naturally upweights observations where the predicted probability is near 0 or 1, where the logit-scale bias is most sensitive.

\subsubsection{Poisson Distribution}

Suppose the outcome $Y \in \{0, 1, 2, \ldots\}$ follows a Poisson distribution.
The canonical link is the log function $h(\mu)= \log(\mu)$, and the cumulant function is $\kappa(\theta)=\exp(\theta)$.
We use the exponential activation in the last layer of the outcome head, yielding $\hat{\mu}(\bX, a) \in \mR^{+}$.
The ADCF specializes to
\begin{align*}
    \psi_a(\mP) = \mathbb{E}\left\{ \log \mathbb{E}[Y \mid \boldsymbol{X}, A = a] \right\},
\end{align*}
i.e., the average log-rate under intervention $do(A=a)$.
The efficient influence function is
\begin{equation*}
    \phi_a(\mP) = \frac{\mathds{1}(A=a)\left( Y - \mu(\bx,a)\right)}{\pi(a \mid \bx)\mu(\bx, a)}  + \log \mu(\bx, a)   - \psi_a,
\end{equation*}
where $h'(\mu) = 1/\mu$ for the log link.

Substituting $\kappa(\theta) = e^\theta$ and $h'(\mu) = 1/\mu$ into~(\ref{tr_term}) yields
\begin{equation*}
    \begin{aligned}
    \mathcal{R}(\mu, \pi, \epsilon) =& \frac{1}{n}\sum_{i=1}^n\left\{-y_i\left[\log\mu(\bx_i, a_i)+\frac{\epsilon(a_i)}{\pi(a_i|\bx_i)\mu(\bx_i, a_i)} \right]\right.
\\+&\left.\exp\left(\log\mu(\bx_i, a_i)+\frac{\epsilon(a_i)}{\pi(a_i|\bx_i)\mu(\bx_i, a_i)}\right)\right\}.
\end{aligned}
\end{equation*}

The argument to $\exp(\cdot)$ is the fluctuated log-rate.
In the Poisson case, the correction term scales inversely with the conditional mean $\mu$, reflecting the fact that the log-scale is more sensitive for small counts.

\begin{remark}[Unifying perspective]
Both the Bernoulli and Poisson examples follow from the same general recipe: (1)~choose the canonical link $h$ and cumulant $\kappa$ for the distribution; (2)~compute $h'(\mu)$; (3)~substitute into~(\ref{tr_term}).
This unified derivation applies to any member of the EDF, including the Gamma, inverse Gaussian, and negative binomial distributions, by simply plugging in the appropriate $\kappa$ and $h$.
The Gaussian case (where $h = \text{id}$ and $\kappa(\theta) = \theta^2/2$) reduces exactly to the targeted regularization of \citet{nie2021vcnet}, confirming that our framework is a strict generalization.
\end{remark}

\section{Experiments}

\subsection{Experimental Setup} 
\paragraph{Data Generation}
Since the true causal effect are not available for real-world data, previous methods \citep{nie2021vcnet, wang2022generalization} use synthetic/semi-synthetic data for empirical evaluation.
Following them, based on one synthetic dataset and two semi-synthetic datasets, News \citep{schwab2020learning} and TCGA \citep{weinstein2013cancer}, we design two distinct treatment settings to verify the effectiveness of our method: binary and continuous treatments.
And for each setting, we assume the outcomes $Y$ follow Bernoulli and Poisson distributions, respectively, serving as representative examples of EDF.
Note that we do not consider the Gaussian distribution here, because our method is equivalent to existing methods \citep{shi2019adapting, nie2021vcnet} in this case.
Due to the space limit, we provide the
details of data generation in the Appendix \ref{app:syn} and \ref{app:semi}.

\begin{table*}[]
\centering
\caption{Results on binary treatment. The MAE with respect to ATE is reported, the best result is in bold.}
\label{tab:binary results}
\scalebox{0.82}{
\begin{tabular}{cccc|ccc}
\hline
& \multicolumn{3}{c|}{Bernoulli} 
& \multicolumn{3}{c}{Poisson}                                                  \\ \hline
        & Simulation        & News       & TCGA 
        & Simulation        & News     & TCGA      \\ \hline
Causal Forest     & $1.5665_{\pm 0.0174}$ & $1.8309_{\pm 0.0269}$ & $1.8316_{\pm 0.1039}$ & $3.0082_{\pm 0.0801}$ & $7.2009_{\pm 0.1794}$ & $12.998_{\pm 1.3429}$   \\
Dragonnet     & $1.9036_{\pm 0.3238}$ & $1.8491_{\pm 1.7251}$  & $1.7854_{\pm 1.5359}$ & $2.7935_{\pm 0.0309}$ & $6.4721_{\pm 1.2242}$ & $7.1624_{\pm 3.0149}$ \\
Dragonnet(adapt)     & $1.2830_{\pm 0.3662}$ & $1.7816_{\pm 1.1453}$ & 
$1.5784_{\pm 1.4081}$ & $2.6909_{\pm 0.0737}$ & $5.1950_{\pm 0.9277}$ & $7.1419_{\pm 2.4739}$ \\
DINA-learner      & $0.9133_{\pm 0.2385}$ & $0.9854_{\pm 0.2985}$ & $0.8652_{\pm 0.3429}$ & $1.8522_{\pm 0.6704}$ & $1.6080_{\pm 0.7515}$ & $2.3710_{\pm 0.6003}$ \\
DR Estimator      & $0.8599_{\pm 0.1742}$ & $0.6917_{\pm 0.2588}$ & $0.0706_{\pm 0.0441}$ & $1.8421_{\pm 0.0344}$ & $1.2423_{\pm 0.2198}$ & $1.2989_{\pm 0.2539}$ \\ \hline
Ours(w/o. t-reg)      & $1.0198_{\pm 0.2340}$ & $1.0481_{\pm 0.2603}$ & $0.9198_{\pm 0.1244}$ & $1.9840_{\pm 0.1245}$ & $2.0461_{\pm 0.3602}$ & $2.8991_{\pm 0.4251}$ \\
Ours      & \pmb{$0.8283_{\pm 0.1791}$} & \pmb{$0.5817_{\pm 0.1762}$} & \pmb{$0.0635_{\pm 0.0446}$} & \pmb{$1.0470_{\pm 0.0252}$} & \pmb{$1.2127_{\pm 0.2045}$} & \pmb{$1.0962_{\pm 0.1653}$} \\
\hline
\end{tabular}
}
\end{table*}

\begin{table*}[!t]
\centering
\caption{Results on continuous treatment. AMSE with respect to ADCF is reported, the best result is in bold.}
\label{tab:continuous results}
\scalebox{0.82}{
\begin{tabular}{cccc|ccc}
\hline
& \multicolumn{3}{c|}{Bernoulli} 
& \multicolumn{3}{c}{Poisson}                                                  \\ \hline
        & Simulation        & News       & TCGA       
        & Simulation        & News     & TCGA      \\ \hline
Causal Forest      & $0.8886_{\pm 0.0178}$ & $0.4789_{\pm 0.5316}$ & $0.9468_{\pm 0.2079}$ & $0.9282_{\pm 0.0900}$ & $15.295_{\pm 4.4294}$ & $4.6775_{\pm 1.2552}$   \\
VCNet   & $0.9862_{\pm 0.9417}$ & $0.4641_{\pm 0.4874}$ & $1.9497_{\pm 1.9862}$ & $0.8394_{\pm 0.0161}$ & $9.9018_{\pm 3.7146}$ & $3.0901_{\pm 1.0139}$ \\
VCNet(adapt)      & $0.8292_{\pm 0.6323}$ & $0.3041_{\pm 0.2579}$ & $1.3134_{\pm 1.6942}$ & $0.7478_{\pm 0.0212}$ & $3.6986_{\pm 
0.6073}$ & $3.0778_{\pm 1.0212}$ \\
DR Estimator      & $0.1504_{\pm 0.1424}$ & $0.1926_{\pm 0.1499}$ & $0.0916_{\pm 0.0455}$ & $0.5014_{\pm 0.0082}$ & \pmb{$2.3014_{\pm 0.6738}$} & $2.6554_{\pm 0.4012}$ \\ \hline
Ours(w/o. t-reg)       & $0.4091_{\pm 0.3286}$ & $0.1992_{\pm 0.1396}$ & $0.8896_{\pm 0.3711}$ & $0.5239_{\pm 0.1546}$ & $3.0159_{\pm 0.8433}$ & $2.9032_{\pm 0.4075}$ \\
Ours      & \pmb{$0.1111_{\pm 0.1103}$} & \pmb{$0.1424_{\pm 0.1057}$} & \pmb{$0.0443_{\pm 0.0328}$} & \pmb{$0.4177_{\pm 0.0239}$} & $2.4547_{\pm 0.5815}$ & \pmb{$2.3570_{\pm 0.2612}$} \\
\hline
\end{tabular}
}
\end{table*}

\paragraph{Baselines}
For \textit{binary treatment}, we compare our method with:
(1) \textbf{Dragonnet} \citep{shi2019adapting}. It designs targeted regularization similar to ours, which is only applicable to Gaussian distribution.
(2)\textbf{Dragonnet(adapt)}. We replace the Mean Squared Error (MSE) loss in Dragonnet with the negative log-likelihood that fits the actual outcome distribution.
(3) \textbf{DINA-learner} \citep{gao2022estimatingheterogeneoustreatmenteffects}. It extends R-learner framework to accommodate outcomes from the exponential family distributions.
For \textit{continuous treatment}, we compare our method with:
(4) \textbf{VCNet} \citep{nie2021vcnet}, which adapts a varying coefficient model to handle continuous treatment, and designs functional targeted regularization similar to ours, which is only applicable to Gaussian distribution.
Similarly, we consider (5)\textbf{VCNet(adapt)} that replaces the MSE loss for outcome $Y$ in VCNet with the negative log-likelihood that fits the actual outcome distribution.
In \textit{both treatment settings}, we also employ (6)\textbf{Causal Forest} \citep{wager2018estimation} as a baseline which does not impose specific restrictions on the treatment type.
(7) \textbf{DR Estimator} as described in Eq. (\ref{onestep})
and (\ref{eq:dr-curve}).
And \textbf{Ours(w/o. t-reg)} is the simplified version of our method that does not include the designed targeted regularization.

\paragraph{Metric}
For \textit{binary treatment}, we evaluate the Mean Absolute Error (MAE) of the Average Treatment Effect (ATE) as $MAE=|\psi-\hat{\psi}|$, where $\psi$ is the true ATE and $\hat{\psi}$ is the estimated ATE.
For \textit{continuous treatment}, we focus on the Average Mean Squared Error (AMSE) of the ADCF, i.e., $AMSE = \int_{\cA}[\hat{\psi}(a)-\psi(a)]^2 p(a) da$, where $p(a)$ is the marginal density of treatments.
For the synthetic dataset, we randomly sample 60\%/20\%/20\% units for training/validation/test.
For the semi-synthetic datasets, we randomly split each data into training (67\%), validation (23\%), and test (10\%).
The validation dataset is used for hyperparameter selection and early-stopping.
Besides, we perform 10 replications for each dataset to report the mean and standard deviation of the metric on test set.

\subsection{Result and Analysis}
\subsubsection{Overall Performance}
Table \ref{tab:binary results} presents the MAE of estimated ATE under binary treatment setting, while Table \ref{tab:continuous results} reports the AMSE of estimated ADCF under continuous treatment setting.
Overall, our proposed method outperforms baselines.
And several key observations can be drawn from the results:
\begin{itemize}
    \item Causal forest, Dragonnet and VCNet exhibit limited performance across all settings, since their loss functions are designed for Gaussian distribution and therefore do not align with the actual outcome distributions.
    \item Dragonnet(adapt) and VCNet(adapt) perform better because they replaced the MSE loss in the original model with the negative log-likelihood corresponding to the distribution.
    For instance, the binary cross-entropy for the Bernoulli distribution ensures better alignment with the target distributions.
    \item Under the binary treatment setting, DINA-learner outperforms other baselines, since it is designed to handle exponential family outcomes based on the R-learner framework.
    However, its partially linear assumption is relatively strong and may not be suitable.
    \item Ours (w/o. t-reg) still performs better than Dragonnet (adapt) and VCNet (adapt), indicating that the targeted regularization with distribution mismatch may degrade model performance.
    \item DR Estimator performs worse than Ours, as the presence of $\hat{\pi}(\cdot)$ in the denominator of some terms causes it unstable in finite samples.
    Nevertheless, it remains highly competitive.This observation is consistent with findings in \citep{shi2019adapting, nie2021vcnet}.
    \item Ours achieves significant improvements across various settings, highlighting the effectiveness of the designed targeted regularization for EDF.
\end{itemize}



\subsubsection{Sensitivity Analysis}
We also investigate the impact of the hyperparameter $\beta$, which controls the strength of targeted regularization.
Due to the space limit, 
the details can be found in Appendix \ref{app:sensitivity}.

\section{Conclusion }
In this work, we address the problem of how to design a NN-based estimator for exponential family outcome.
Specifically, we first derive the von-Mises expansion of ADCF to show the first-order bias term in plug-in estimator, then we construct a doubly robust estimator by subtracting the estimated bias term and analyze its asymptotic properties.
Leveraging our theoretical findings, we develop a NN-based estimator by generalizing functional targeted regularization to exponential families and give the theoretical convergence rates.
Extensive experimental results verify the correctness of our theory and the effectiveness of our model.

\bibliography{references}

\clearpage

\onecolumn

\title{Targeted Regularization for Causal Effect Estimation with Exponential Dispersion Family Outcomes (Supplementary Material)}
\maketitle

\appendix

\section{Related Works}

\paragraph{NN-based Treatment Effect Estimator}
Nowadays, NNs have emerged as a pivotal tool for treatment effect estimation due to their flexibility and widespread adoption.
Much of the work in this area has focused on mitigating confounding bias through balanced representation learning.
\citep{shalit2017estimating} first give a generalization bound consisting of an empirical loss and an Integral Probability Metric, thus establishing the paradigm of the balanced representation learning.
Building on this, recent studies have incorporated the weighting strategies as an additional correction for confounding bias into this framework, such as \citep{johansson2018learning, hassanpour2019counterfactual}.
\citep{wang2022generalization} and \citep{kazemi2024adversarially} further extend this framework from binary treatment to continuous treatment scenarios.
In parallel, another prominent paradigm has been proposed for treatment effect estimation using NNs, focusing on exploiting the sufficiency of propensity score.
Based on this, \citep{shi2019adapting, nie2021vcnet} introduce targeted regularizations to correct the bias.
Beyond standard feedforward NNs, specialized architectures have also been explored, including Variational Autoencoder (VAE) \citep{louizos2017causal}, Generative Adversarial Network (GAN) \citep{yoon2018ganite}.
Compared to the above methods, our method is not limited to Gaussian-distributed outcome and equipped with theoretical guarantees for asymptotic correctness, addressing key limitations of prior works.

\paragraph{DML, TMLE and Targeted Regularization}
To address the first-order bias in the plug-in estimator, the Doubly Machine Learning (DML), which is first proposed by \citep{vanderLaan+2006} and further developed by \citep{Chernozhukov2017, Chernozhukov2018}—achieves doubly robustness by incorporating a bias correction term.
Their theoretical work demonstrated that these doubly robust estimators attain $\sqrt{n}$-convergence rates under appropriate conditions.
Following the DML framework, \citep{NieRLearner} introduced the R-Learner, a general class of two-step algorithms for estimating treatment effects in observational studies.
\citep{gao2022estimatingheterogeneoustreatmenteffects} extended the R-Learner framework to accommodate exponential family outcomes and introduced DINA (Difference in Natural pArameters) as a measure of treatment effects.
Targeted Maximum Likelihood Estimation (TMLE) \citep{van2011targeted}, targeted regularization \citep{shi2019adapting}, and functional targeted regularization \citep{nie2021vcnet} offer an alternative framework to one-step correction by correcting bias on the distributional scale.
Similarly, \citep{vanderlaan2024combiningtlearningdrlearningframework} and \citep{vansteelandt2023orthogonalpredictioncounterfactualoutcomes} employ sieve methods to debias plug-in estimators using the TMLE framework.
And \citep{kennedy2023semiparametricdoublyrobusttargeted} provides a comprehensive review of doubly robustness from a semiparametric perspective, with particular emphasis on minimax-style efficiency bounds, detailed worked examples, and practical derivation shortcuts.
To the best of our knowledge, the only prior work on exponential family outcomes \citep{gao2022estimatingheterogeneoustreatmenteffects} only focuses on binary treatment and is limited to the partially linear assumption.
Meanwhile, prior works applying targeted regularization to NNs \citep{shi2019adapting, nie2021vcnet} have been limited to the Gaussian-distributed outcomes setting.
Different from them, our work generalizes targeted functional regularization for exponential family outcomes, extending the framework to both binary and continuous treatment regimes.

\section{Experiment Details}
\subsection{Synthetic Data Generation}\label{app:syn}
We simulate the synthetic dataset as follows.
We first generate 10000 samples with covariates $\bX \sim \text{Unif}(0,1) \in \mathbb{R}^6$, and the treatments and outcomes under different settings are generated as follows:
\begin{itemize}
    \item  To generate the treatments, we set:
    \begin{align*}
        A &= \begin{cases} 
            \cB(1, \sigma(\tilde{a})), & \text{for binary case}, \\
            \sigma(\tilde{a})\;\;\;\;\;\;\;\;\;\;\;\;\;, & \text{for continuous case} ,
        \end{cases}
    \end{align*}
    where $\tilde{a} = 10 \frac{\sin\left(\max(\bX_1, \bX_2, \bX_3)\right) + \max(\bX_3, \bX_4, \bX_5)^3}{1 + (\bX_1 + \bX_5)^2} + \sin(0.5\bX_3)(1 + \exp(\bX_4 - 0.5\bX_3)) + \bX_3^2 + 2\sin(\bX_4) + 2\bX_5 - 6.5 + \mathcal{N}(\mu_1, 0.5)$, $\cB(1, p)$ denotes the Bernoulli distribution with probability $p$, and $\sigma(\cdot)$ denotes the sigmoid function.
    \item After obtaining the assigned treatments, we generate the corresponding outcomes under different distributions.
    Specifically, we first generate $\tilde{\mu}=2(A+\gamma)\sin(\bX_4)(A+4\max(\bX_1, \bX_6)^3) /(1+2\bX_3^2) $, where $\gamma$ is set to -0.5 for the Bernoulli distribution and 0.5 for the Poisson distribution, then the corresponding outcomes are obtained as follows:
    \begin{align*}
       Y = 
    \begin{cases} 
        \cB(1, \sigma(\tilde{\mu})), & \text{for Bernoulli case}, \\
        \cP(\exp(clip(\tilde{\mu}, -4, 4))), & \text{for Poisson case} ,
    \end{cases} 
    \end{align*}
    where $\cP(\lambda)$ denotes the Poisson distribution with parameter $\lambda$, and we limit $\tilde{\mu}$ to the range of -4 to 4 to avoid extreme values after exponential scaling.
\end{itemize}

\subsection{Semi-synthetic Data Generation} \label{app:semi}
Following \cite{schwab2020learning, bica2020estimating, nie2021vcnet},
we reuse the covariates $X$ of the real-word datasets, News and TCGA.
To generate the assigned treatments and their corresponding outcomes,
we first generate a set of parameters $\boldsymbol{V}_{i}=\boldsymbol{U}_{i}/||\boldsymbol{U}_{i}||$ and $i=1,2,3$, where $\boldsymbol{U}_{i}$ is sampled from a normal distribution $\mathcal{N}(\mathbf{0}, \mathbf{1})$,
then:
\begin{itemize}
    \item To assign the treatments, we generate $\tilde{a}=\left| w *\frac{\mathbf{V}_3^\top \bX}{\mathbf{V}_2^\top \bX} \right|$,
    where the parameter $w$ varies depending on the dataset and treatment type.
    For the News dataset, we set $w=1.5$ for binary treatments and $w=0.5$ for continuous treatments.
    For the TCGA dataset, $w$ is set to 5 for binary treatments and 0.2 for continuous treatments. 
    Using these values, 
    the treatments are generated as:
    \begin{align*}
        A &= \begin{cases} 
            \cB(1, \sigma(\tilde{a}))\;\;, & \text{for binary case}, \\
            Beta(2, |\tilde{a}|), & \text{for continuous case} .
        \end{cases}
    \end{align*}

    \item To generate the outcomes, we first generate $\tilde{\mu}$ according to the treatments setting.
    For binary treatments, we set $\tilde{\mu}=\cos\left(1.2\pi A\right) \times 2(\max(-2, \frac{\mathbf{V}_2^\top \bX}{\mathbf{V}_3^\top \bX + 2} - 0.3)) + 10 \mathbf{V}_1^\top \bX$.
    For continuous treatments, $\tilde{\mu}$ is given by $\tilde{\mu}=20(A-0.5)\sin{(\pi A)}\left(\max(\alpha, \frac{\mathbf{V}_2^\top \bX}{\mathbf{V}_3^\top \bX + 2} - 0.3)+\beta \mathbf{V}_1^\top \bX\right)$.
    Based on the computed $\tilde{\mu}$,
    the outcomes $Y$ are generated according to the following rules:
    \begin{align*}
        Y = \begin{cases} 
            \cB(1, \sigma(\mu)), & \text{for Bernoulli case}, \\
            \cP(\exp(\max(4, \gamma*\mu))), & \text{for Poisson case} .
        \end{cases}
    \end{align*}

    In particular, we set $\alpha=-2,\beta=10,\gamma=2.5$ for the News dataset and $\alpha=-0.5,\beta=5,\gamma=4.5$ for the TCGA dataset.
\end{itemize}

\subsection{Sensitivity Analysis} \label{app:sensitivity}
\begin{figure}[!t]
    \centering
    \subfigure[Bernoulli]{\includegraphics[width=.35\linewidth]{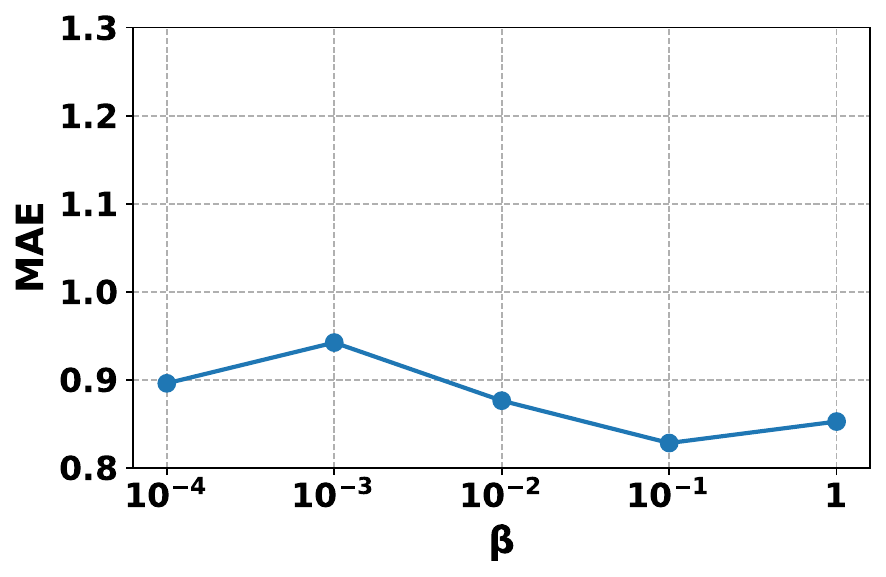}}
    \subfigure[Poisson]{\includegraphics[width=.35\linewidth]{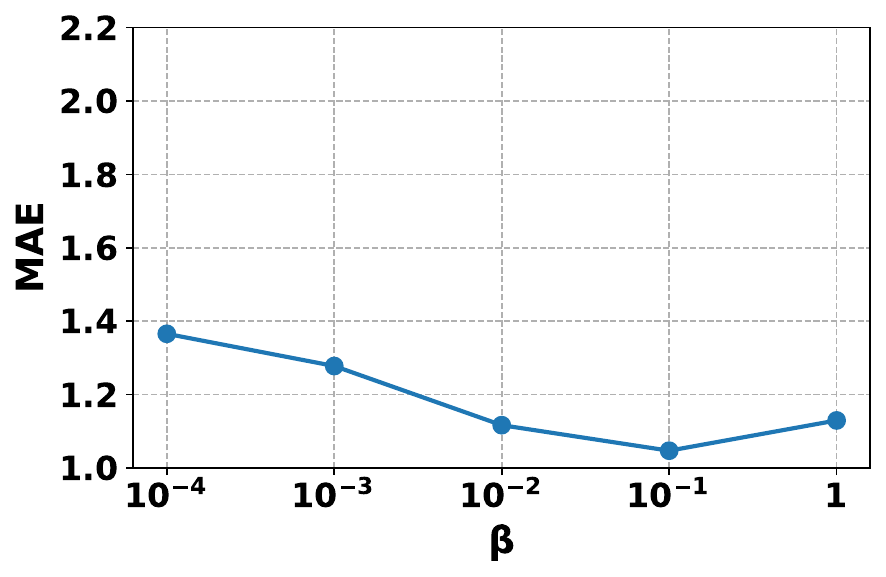}}

    \caption{Sensitivity analysis on synthetic data under binary treatment setting.}
    \label{fig:sensitivity}
\end{figure}

We take synthetic dataset under binary treatment setting as an example to evaluate the sensitivity of the model to the parameter $\beta$, which controls the strength of the targeted regularization.
By varying the values of $\beta$, we plot the results in Fig. (\ref{fig:sensitivity}).
We observe that the MAE slightly increase when $\beta$ becomes either too small or too large.
Specifically, when $\beta$ is set too small, it fails to correct the confounding bias.
Conversely, when $\beta$ is set too large, it interferes with the estimation of nuisance-function estimators, which are crucial for constructing the targeted regularization term.
However, although varying $\beta$ does affect the model's performance to some extent, the model still performs competitively compared to the baselines.

\section{Proof Details}
We maintain the numbering and
notation of the main text throughout.  We denote observations by
$\bZ=(Y,A,\bX)$, the conditional mean by
$\mu(\bx,a)=\mE[Y\mid\bX=\bx,\,A=a]$, the generalised propensity
score by $\pi(a\mid\bx)$, and the functional of interest by
$\psi_a(\mP)=\mE\bigl\{h\bigl(\mu(\bX,a)\bigr)\bigr\}$ for a twice
continuously differentiable link function~$h$.  Where a second
probability measure $\bar{\mP}$ appears, we write $\bar\mu$ and
$\bar\pi$ for the corresponding conditional mean and propensity, and
$\mP_n=n^{-1}\sum_{i=1}^{n}\delta_{\bZ_i}$ for the empirical measure.

Throughout, all norms $\lVert\cdot\rVert$ without subscript denote the
$L^2(\mP)$-norm unless stated otherwise.

\subsection{Proof of Lemma \ref{vonmiseexpansion} (Von Mises expansion for discrete
treatment)}
\label{sec:proof-lemma41}


\begin{assumption}[Standing regularity for Lemma~4.1 and
Corollary~4.2]
\begin{enumerate}
    \item[\textup{(R1)}]
    \textbf{Overlap.}  There exists $c_\pi>0$ such that
    $\pi(a\mid\bx)\geq c_\pi$ for all $a\in\cA$, $\bx\in\cX$.
    When $\bar\mP$ is involved, additionally
    $\bar\pi(a\mid\bx)\geq c_\pi$.
    \item[\textup{(R2)}]
    \textbf{Bounded link derivatives.}  There exist constants
    $M_1,M_2<\infty$ such that $|h'(t)|\leq M_1$ and
    $|h''(t)|\leq M_2$ for every $t$ in the range of $\mu$
    (and of $\bar\mu$).
    \item[\textup{(R3)}]
    \textbf{Bounded outcome.}  $\mE[Y^2\mid\bX,A]$ is uniformly
    bounded, and $\mu$ (and $\bar\mu$) are uniformly bounded.
\end{enumerate}
\label{as:standing-41}
\end{assumption}

\begin{lemma}[Von Mises expansion for discrete treatment]%
\label{lem:vonMises-discrete}
Let $\psi_a(\mP)=\mE\bigl\{h\bigl(\mE[Y\mid\bX,\,A=a]\bigr)\bigr\}$
with $h$ twice continuously differentiable.  Under
Assumption~\ref{as:standing-41}, for another probability measure
$\bar{\mP}$, the functional $\psi_a$ admits the von Mises expansion
\begin{equation}\label{eq:vonMises}
    \psi_a(\bar{\mP})-\psi_a(\mP)
    \;=\;
    \int\phi_a(\bz;\bar{\mP})\,d(\bar{\mP}-\mP)(\bz)
    \;+\;R_2(\bar{\mP},\mP),
\end{equation}
where the \emph{efficient influence function} is
\begin{equation}\label{eq:EIF}
    \phi_a(\bz;\mP)
    \;=\;
    \frac{\mathbbm{1}(A=a)}{\pi(a\mid\bx)}\,
    \bigl\{Y-\mu(\bx,a)\bigr\}\,h'\!\bigl(\mu(\bx,a)\bigr)
    \;+\;h\!\bigl(\mu(\bx,a)\bigr)-\psi_a(\mP),
\end{equation}
and the second-order remainder is
\begin{equation}\label{eq:R2}
\begin{aligned}
    R_2(\bar{\mP},\mP)
    \;&=\;
    \frac{1}{2}\int h''\!\bigl(\mu^*(\bx,a)\bigr)\,
    \bigl[\bar\mu(\bx,a)-\mu(\bx,a)\bigr]^2\,d\mP(\bx)\\
    &\quad+\;\int\!\left\{
    \frac{\pi(a\mid\bx)}{\bar\pi(a\mid\bx)}-1\right\}
    h'\!\bigl(\bar\mu(\bx,a)\bigr)\,
    \bigl(\mu(\bx,a)-\bar\mu(\bx,a)\bigr)\,d\mP(\bx),
\end{aligned}
\end{equation}
with $\mu^*(\bx,a)$ an intermediate value between $\mu(\bx,a)$
and $\bar\mu(\bx,a)$ (depending on~$\bx$), given by Taylor's
theorem.
\end{lemma}

\begin{proof}
The proof has two parts: (A)~derivation of the efficient influence
function via the Gateaux-derivative approach of
\citet{kennedy2023semiparametricdoublyrobusttargeted}; (B)~derivation
of the second-order remainder.

\medskip
\noindent\textbf{Part A.  Derivation of the influence function.}

We compute the Gateaux derivative of $\psi_a$ at $\mP$ in the
direction of a Dirac measure.  We present the calculation assuming a
discrete data-generating distribution; the resulting influence
function~\eqref{eq:EIF} depends on $\bz$ only through the
regression functions $\mu(\bx,a)$ and $\pi(a\mid\bx)$ and is
therefore well-defined---and serves as the efficient influence
function---in discrete, continuous, and mixed settings alike, as
shown by the general semiparametric theory
\citep[Section~2.3]{kennedy2023semiparametricdoublyrobusttargeted}.

Fix an arbitrary point $\bz'=(y',a',\bx')$ and define the
contaminated mass function
\[
    p_\epsilon(\bz)
    \;=\;(1-\epsilon)\,p(\bz)+\epsilon\,\mathbbm{1}(\bz=\bz'),
    \qquad \epsilon\in[0,1).
\]
The Gateaux derivative of $\psi_a$ at~$\mP$ in the direction
$\delta_{\bz'}-\mP$ is
\begin{equation}\label{eq:gateaux-def}
    \phi_a(\bz';\mP)
    \;=\;
    \frac{\partial}{\partial\epsilon}\,
    \psi_a\!\bigl(p_\epsilon\bigr)\bigg|_{\epsilon=0}.
\end{equation}

\noindent\emph{Step A.1.  Submodel components.}
From $p_\epsilon$ we extract
\begin{align}
    p_\epsilon(\bx)
    &=(1-\epsilon)\,p(\bx)+\epsilon\,\mathbbm{1}(\bx=\bx'),
    \label{eq:sub-margX}\\[4pt]
    p_\epsilon(a\mid\bx)
    &=\frac{(1-\epsilon)\,p(a,\bx)
    +\epsilon\,\mathbbm{1}(a=a',\,\bx=\bx')}%
    {(1-\epsilon)\,p(\bx)
    +\epsilon\,\mathbbm{1}(\bx=\bx')},
    \label{eq:sub-propensity}\\[4pt]
    p_\epsilon(y\mid\bx,a)
    &=\frac{(1-\epsilon)\,p(\bz)
    +\epsilon\,\mathbbm{1}(\bz=\bz')}%
    {(1-\epsilon)\,p(a,\bx)
    +\epsilon\,\mathbbm{1}(a=a',\,\bx=\bx')}.
    \label{eq:sub-outcome}
\end{align}

\noindent\emph{Step A.2.  Derivative of the conditional outcome
density.}
Differentiating~\eqref{eq:sub-outcome} as a ratio at $\epsilon=0$
(the denominator is $p(a,\bx)>0$ by the overlap
Assumption~\ref{as:standing-41}(R1)):
\begin{align}
    \frac{\partial}{\partial\epsilon}\,
    p_\epsilon(y\mid\bx,a)\bigg|_{\epsilon=0}
    &=\frac{\mathbbm{1}(\bz=\bz')-p(\bz)}{p(a,\bx)}
    -p(y\mid\bx,a)\,
    \frac{\mathbbm{1}(a=a',\bx=\bx')-p(a,\bx)}{p(a,\bx)}
    \notag\\
    &=\mathbbm{1}(a=a',\,\bx=\bx')\,
    \frac{\mathbbm{1}(y=y')-p(y\mid\bx,a)}{p(a,\bx)}.
    \label{eq:deriv-cond-outcome}
\end{align}
The simplification uses
$\mathbbm{1}(\bz=\bz')=\mathbbm{1}(y=y')\,
\mathbbm{1}(a=a',\bx=\bx')$ and collects the common factor
$\mathbbm{1}(a=a',\bx=\bx')/p(a,\bx)$.

\noindent\emph{Step A.3.  Derivative of $\psi_a$.}
Write $\mu_\epsilon(\bx,a)=\sum_{y}y\,p_\epsilon(y\mid\bx,a)$ and
$\psi_a(p_\epsilon)=\sum_{\bx}h\!\bigl(\mu_\epsilon(\bx,a)\bigr)\,
p_\epsilon(\bx)$.  By the product rule:
\begin{equation}\label{eq:product-rule}
    \frac{\partial}{\partial\epsilon}\,\psi_a(p_\epsilon)
    \bigg|_{\epsilon=0}
    =\underbrace{\sum_{\bx}h'\!\bigl(\mu(\bx,a)\bigr)\,
    \dot\mu(\bx,a)\,p(\bx)}_{\displaystyle(T_1)}
    \;+\;
    \underbrace{\sum_{\bx}h\!\bigl(\mu(\bx,a)\bigr)\,
    \dot{p}(\bx)}_{\displaystyle(T_2)},
\end{equation}
where $\dot\mu(\bx,a)
=\frac{\partial}{\partial\epsilon}\mu_\epsilon(\bx,a)|_{\epsilon=0}$
and $\dot{p}(\bx)
=\frac{\partial}{\partial\epsilon}p_\epsilon(\bx)|_{\epsilon=0}$.

From~\eqref{eq:sub-margX},
$\dot{p}(\bx)=\mathbbm{1}(\bx=\bx')-p(\bx)$.
From~\eqref{eq:deriv-cond-outcome},
\[
    \dot\mu(\bx,a)
    =\sum_{y}y\,\mathbbm{1}(a=a',\bx=\bx')\,
    \frac{\mathbbm{1}(y=y')-p(y\mid\bx,a)}{p(a,\bx)}
    =\mathbbm{1}(a=a',\bx=\bx')\,
    \frac{y'-\mu(\bx,a)}{p(a,\bx)}.
\]
Substituting into $(T_1)$:
\begin{align*}
    (T_1)
    &=\sum_{\bx}h'\!\bigl(\mu(\bx,a)\bigr)\,
    \mathbbm{1}(a=a',\bx=\bx')\,
    \frac{y'-\mu(\bx,a)}{p(a,\bx)}\,p(\bx)\\
    &=\frac{\mathbbm{1}(a'=a)}{\pi(a\mid\bx')}\,
    \bigl\{y'-\mu(\bx',a)\bigr\}\,h'\!\bigl(\mu(\bx',a)\bigr),
\end{align*}
using $p(a,\bx')/p(\bx')=\pi(a\mid\bx')$ and the sum collapsing on
$\bx=\bx'$.  For $(T_2)$:
\[
    (T_2)=h\!\bigl(\mu(\bx',a)\bigr)-\sum_{\bx}h\!\bigl(\mu(\bx,a)
    \bigr)\,p(\bx)
    =h\!\bigl(\mu(\bx',a)\bigr)-\psi_a(\mP).
\]
Combining $(T_1)$ and $(T_2)$ and replacing the dummy point
$\bz'=(y',a',\bx')$ by $\bz=(Y,A,\bX)$:
\begin{equation}\label{eq:IF-derived}
    \phi_a(\bz;\mP)
    =\frac{\mathbbm{1}(A=a)}{\pi(a\mid\bx)}\,
    \bigl\{Y-\mu(\bx,a)\bigr\}\,h'\!\bigl(\mu(\bx,a)\bigr)
    +h\!\bigl(\mu(\bx,a)\bigr)-\psi_a(\mP),
\end{equation}
which is~\eqref{eq:EIF}.

\medskip
\noindent\textbf{Part B.  Derivation of the second-order remainder.}

By definition, the remainder in the von Mises expansion is
\[
    R_2(\bar\mP,\mP)
    =\psi_a(\bar\mP)-\psi_a(\mP)
    -\int\phi_a(\bz;\bar\mP)\,d(\bar\mP-\mP)(\bz).
\]
Since the influence function has mean zero under its own measure,
$\mE_{\bar\mP}[\phi_a(\bZ;\bar\mP)]=0$
\citep[Lemma~2.2]{kennedy2023semiparametricdoublyrobusttargeted},
the integral simplifies:
\begin{equation}\label{eq:R2-def-simplified}
    R_2(\bar\mP,\mP)
    =\psi_a(\bar\mP)-\psi_a(\mP)
    +\mE_{\mP}\!\bigl[\phi_a(\bZ;\bar\mP)\bigr].
\end{equation}

\noindent\emph{Step B.1.  Expanding $\mE_\mP[\phi_a(\bZ;\bar\mP)]$.}
Substituting the form~\eqref{eq:EIF} evaluated under $\bar\mP$:
\begin{align}
    \mE_{\mP}\!\bigl[\phi_a(\bZ;\bar\mP)\bigr]
    &=\mE_{\mP}\!\left[
    \frac{\mathbbm{1}(A=a)}{\bar\pi(a\mid\bX)}\,
    \bigl\{Y-\bar\mu(\bX,a)\bigr\}\,
    h'\!\bigl(\bar\mu(\bX,a)\bigr)\right]
    +\mE_{\mP}\!\bigl[h\!\bigl(\bar\mu(\bX,a)\bigr)\bigr]
    -\psi_a(\bar\mP). \label{eq:E-phi-bar}
\end{align}
For the first term, we apply the tower property of conditional
expectation, conditioning on $(\bX,A)$ under~$\mP$.
The indicator $\mathbbm{1}(A=a)$ restricts to $A=a$, so
\begin{align*}
    \mE_{\mP}\!\left[
    \frac{\mathbbm{1}(A=a)}{\bar\pi(a\mid\bX)}\,
    \bigl\{Y-\bar\mu(\bX,a)\bigr\}\,
    h'\!\bigl(\bar\mu(\bX,a)\bigr)\right]
    &=\int_{\cX}\frac{\pi(a\mid\bx)}{\bar\pi(a\mid\bx)}\,
    \mE_{\mP}\!\bigl[Y-\bar\mu(\bx,a)\mid\bX=\bx,A=a\bigr]\,
    h'\!\bigl(\bar\mu(\bx,a)\bigr)\,d\mP(\bx)\\
    &=\int_{\cX}\frac{\pi(a\mid\bx)}{\bar\pi(a\mid\bx)}\,
    \bigl(\mu(\bx,a)-\bar\mu(\bx,a)\bigr)\,
    h'\!\bigl(\bar\mu(\bx,a)\bigr)\,d\mP(\bx),
\end{align*}
since $\mE_\mP[Y\mid\bX=\bx,A=a]=\mu(\bx,a)$.

Substituting back into~\eqref{eq:R2-def-simplified} and using
$\psi_a(\bar\mP)=\mE_{\bar\mP}[h(\bar\mu(\bX,a))]$,
$\psi_a(\mP)=\mE_\mP[h(\mu(\bX,a))]$:
\begin{equation}\label{eq:R2-intermediate}
\begin{aligned}
    R_2(\bar\mP,\mP)
    &=\int_{\cX}\frac{\pi(a\mid\bx)}{\bar\pi(a\mid\bx)}\,
    h'\!\bigl(\bar\mu(\bx,a)\bigr)\,
    \bigl(\mu(\bx,a)-\bar\mu(\bx,a)\bigr)\,d\mP(\bx)\\
    &\quad+\int_{\cX}\bigl[
    h\!\bigl(\bar\mu(\bx,a)\bigr)-h\!\bigl(\mu(\bx,a)\bigr)
    \bigr]\,d\mP(\bx).
\end{aligned}
\end{equation}
Here the cross-measure terms involving $\psi_a(\bar\mP)$ cancel
exactly with the $-\psi_a(\bar\mP)$ from~\eqref{eq:E-phi-bar}.

\noindent\emph{Step B.2.  Taylor expansion.}
Since $h$ is twice continuously differentiable, Taylor's theorem with
integral (Lagrange) remainder gives, for each~$\bx$,
\begin{equation}\label{eq:taylor}
    h\!\bigl(\mu(\bx,a)\bigr)-h\!\bigl(\bar\mu(\bx,a)\bigr)
    =h'\!\bigl(\bar\mu(\bx,a)\bigr)\,
    \bigl(\mu(\bx,a)-\bar\mu(\bx,a)\bigr)
    +\tfrac{1}{2}\,h''\!\bigl(\mu^*(\bx,a)\bigr)\,
    \bigl(\mu(\bx,a)-\bar\mu(\bx,a)\bigr)^2,
\end{equation}
where $\mu^*(\bx,a)$ lies between $\mu(\bx,a)$ and $\bar\mu(\bx,a)$
(depending on~$\bx$), by the mean-value form of the remainder.

Rearranging~\eqref{eq:taylor}:
\[
    h\!\bigl(\bar\mu(\bx,a)\bigr)-h\!\bigl(\mu(\bx,a)\bigr)
    =-h'\!\bigl(\bar\mu(\bx,a)\bigr)\,
    \bigl(\mu(\bx,a)-\bar\mu(\bx,a)\bigr)
    -\tfrac{1}{2}\,h''\!\bigl(\mu^*(\bx,a)\bigr)\,
    \bigl(\mu(\bx,a)-\bar\mu(\bx,a)\bigr)^2.
\]
Substituting into~\eqref{eq:R2-intermediate}:
\begin{align*}
    R_2(\bar\mP,\mP)
    &=\int_{\cX}\frac{\pi(a\mid\bx)}{\bar\pi(a\mid\bx)}\,
    h'\!\bigl(\bar\mu(\bx,a)\bigr)\,
    \bigl(\mu(\bx,a)-\bar\mu(\bx,a)\bigr)\,d\mP(\bx)\\
    &\quad-\int_{\cX}h'\!\bigl(\bar\mu(\bx,a)\bigr)\,
    \bigl(\mu(\bx,a)-\bar\mu(\bx,a)\bigr)\,d\mP(\bx)\\
    &\quad-\tfrac{1}{2}\int_{\cX}h''\!\bigl(\mu^*(\bx,a)\bigr)\,
    \bigl(\mu(\bx,a)-\bar\mu(\bx,a)\bigr)^2\,d\mP(\bx)\\[4pt]
    &=\int_{\cX}\left\{
    \frac{\pi(a\mid\bx)}{\bar\pi(a\mid\bx)}-1\right\}
    h'\!\bigl(\bar\mu(\bx,a)\bigr)\,
    \bigl(\mu(\bx,a)-\bar\mu(\bx,a)\bigr)\,d\mP(\bx)\\
    &\quad+\tfrac{1}{2}\int_{\cX}h''\!\bigl(\mu^*(\bx,a)\bigr)\,
    \bigl(\bar\mu(\bx,a)-\mu(\bx,a)\bigr)^2\,d\mP(\bx),
\end{align*}
where in the last line we used
$(\mu-\bar\mu)^2=(\bar\mu-\mu)^2$ and reversed the sign convention
to match~\eqref{eq:R2}.  This is exactly~\eqref{eq:R2}.
\end{proof}

\subsection{Proof of Corollary \ref{asydistr} (Asymptotic distribution of the
DR estimator)}
\label{sec:proof-cor42}

\begin{corollary}\label{cor:asy}
Under Assumption~\ref{as:standing-41}, if the following conditions hold:
\begin{enumerate}
    \item[\textup{(i)}]
    $(\mP_n-\mP)\bigl\{\phi_a(\hat{\mP})-\phi_a(\mP)\bigr\}
    =o_{\mP}(n^{-1/2})$;
    \item[\textup{(ii)}]
    $\lVert\hat\pi(a\mid\cdot)-\pi(a\mid\cdot)\rVert
    =o_{\mP}(n^{-1/4})$;
    \item[\textup{(iii)}]
    $\lVert\hat\mu(\cdot,a)-\mu(\cdot,a)\rVert
    =o_{\mP}(n^{-1/4})$;
\end{enumerate}
then
$\hat\psi_a^{\textup{dr}}-\psi_a(\mP)
=(\mP_n-\mP)\bigl\{\phi_a(\mP)\bigr\}+o_{\mP}(n^{-1/2})$,
and consequently $\hat\psi_a^{\textup{dr}}$ is $\sqrt{n}$-consistent
and asymptotically normal:
\[
    \sqrt{n}\,\bigl(\hat\psi_a^{\textup{dr}}-\psi_a(\mP)\bigr)
    \;\xrightarrow{d}\;
    \cN\!\bigl(0,\;\operatorname{Var}_{\mP}[\phi_a(\bZ;\mP)]\bigr).
\]
\end{corollary}

\begin{proof}

\noindent\textbf{Step 1.  Three-term decomposition.}
The doubly robust estimator is
$\hat\psi_a^{\text{dr}}=\mP_n[\phi_a(\bZ;\hat\mP)]
+\psi_a(\hat\mP)$.
By the von Mises expansion (Lemma~\ref{lem:vonMises-discrete}) with
$\bar\mP=\hat\mP$:
\begin{equation}\label{eq:dr-decomp}
    \hat\psi_a^{\text{dr}}-\psi_a(\mP)
    =\underbrace{(\mP_n-\mP)\bigl[\phi_a(\mP)\bigr]}_{(A)}
    +\underbrace{(\mP_n-\mP)\bigl[\phi_a(\hat\mP)
    -\phi_a(\mP)\bigr]}_{(B)}
    +\underbrace{R_2(\hat\mP,\mP)}_{(C)}.
\end{equation}
Term~$(A)$ is the leading term.  We show $(B)$ and $(C)$ are each
$o_\mP(n^{-1/2})$.

\medskip
\noindent\textbf{Step 2.  Term~$(B)$ is $o_\mP(n^{-1/2})$.}
This is precisely condition~(i).

\medskip
\noindent\textbf{Step 3.  Term~$(C)$ is $o_\mP(n^{-1/2})$.}
From~\eqref{eq:R2} with $\bar\mP=\hat\mP$:
\[
    R_2(\hat\mP,\mP)
    =\int_{\cX}\!\left\{
    \frac{\pi(a\mid\bx)}{\hat\pi(a\mid\bx)}-1\right\}
    h'\!\bigl(\hat\mu(\bx,a)\bigr)\,
    \bigl(\mu(\bx,a)-\hat\mu(\bx,a)\bigr)\,d\mP(\bx)
    +\tfrac12\!\int_{\cX}h''\!\bigl(\mu^*(\bx,a)\bigr)\,
    \bigl(\hat\mu(\bx,a)-\mu(\bx,a)\bigr)^2d\mP(\bx).
\]
For the first integral, note that
\[
    \frac{\pi(a\mid\bx)}{\hat\pi(a\mid\bx)}-1
    =\frac{\pi(a\mid\bx)-\hat\pi(a\mid\bx)}{\hat\pi(a\mid\bx)}.
\]
By Assumption~\ref{as:standing-41}(R1),
$\hat\pi(a\mid\bx)\geq c_\pi$, and by~(R2),
$|h'|\leq M_1$, $|h''|\leq M_2$.  Hence
\begin{align*}
    |R_2(\hat\mP,\mP)|
    &\leq\frac{M_1}{c_\pi}\int_{\cX}
    \bigl|\pi(a\mid\bx)-\hat\pi(a\mid\bx)\bigr|\;
    \bigl|\mu(\bx,a)-\hat\mu(\bx,a)\bigr|\,d\mP(\bx)
    +\frac{M_2}{2}\int_{\cX}
    \bigl(\hat\mu(\bx,a)-\mu(\bx,a)\bigr)^2\,d\mP(\bx).
\end{align*}
By the Cauchy--Schwarz inequality applied to the first integral:
\begin{equation}\label{eq:R2-bound-CS}
    |R_2(\hat\mP,\mP)|
    \leq\frac{M_1}{c_\pi}\,
    \lVert\hat\pi(a\mid\cdot)-\pi(a\mid\cdot)\rVert\;
    \lVert\hat\mu(\cdot,a)-\mu(\cdot,a)\rVert
    +\frac{M_2}{2}\,
    \lVert\hat\mu(\cdot,a)-\mu(\cdot,a)\rVert^2.
\end{equation}
Under conditions~(ii) and~(iii):
\[
    |R_2(\hat\mP,\mP)|
    =O_\mP\!\bigl(o(n^{-1/4})\cdot o(n^{-1/4})\bigr)
    +O_\mP\!\bigl(o(n^{-1/4})^2\bigr)
    =o_\mP(n^{-1/2}).
\]

\medskip
\noindent\textbf{Step 4.  Conclusion.}
Combining Steps~1--3:
\[
    \hat\psi_a^{\text{dr}}-\psi_a(\mP)
    =(\mP_n-\mP)\bigl[\phi_a(\mP)\bigr]+o_\mP(n^{-1/2}).
\]
Since $\phi_a(\bZ;\mP)$ has mean zero ($\mE_\mP[\phi_a]=0$,
Lemma~\ref{lem:vonMises-discrete}) and finite variance
($\operatorname{Var}_\mP[\phi_a]<\infty$ by
Assumption~\ref{as:standing-41}(R1)--(R3)), the classical CLT gives
\[
    \sqrt{n}\;(\mP_n-\mP)\bigl[\phi_a(\mP)\bigr]
    \;\xrightarrow{d}\;
    \cN\!\bigl(0,\operatorname{Var}_\mP[\phi_a(\bZ;\mP)]\bigr).
\]
The remainder $o_\mP(n^{-1/2})$ is $o_\mP(1)$ after scaling by
$\sqrt{n}$, so it vanishes in distribution.  Formally, for any
$\epsilon>0$,
$\mP\!\bigl(\bigl|\sqrt{n}\,o_\mP(n^{-1/2})\bigr|>\epsilon\bigr)
\to 0$,
and the conclusion follows by Slutsky's theorem
\citep[Theorem~2.8]{vandervaart1998asymptotic}:
\[
    \sqrt{n}\,\bigl(\hat\psi_a^{\text{dr}}-\psi_a(\mP)\bigr)
    =\sqrt{n}\;(\mP_n-\mP)[\phi_a(\mP)]
    +\sqrt{n}\cdot o_\mP(n^{-1/2})
    \;\xrightarrow{d}\;
    \cN\!\bigl(0,\operatorname{Var}_\mP[\phi_a(\bZ;\mP)]\bigr).
    \qedhere
\]
\end{proof}

\subsection{Proof of Lemma \ref{lem:sieve-VonMises} (Von Mises expansion for sieve-projected
ADCF)}
\label{sec:proof-lemma43}

For reference, we restate the relevant definitions.  Under
Assumption~4.3 (regularity for continuous treatments), the B-spline
basis is $\bB(a)=(B_1(a),\ldots,B_{K_n}(a))^\top$, the Gram matrix
is $G=\int_{\cA}\bB(a)\bB(a)^\top da$, the sieve coefficient
functional is
$\gamma_k(\mP)=\int_{\cA}\psi_a(\mP)\,B_k(a)\,da$, and the
sieve-projected ADCF is
$\psi^s_a(\mP)=\bB(a)^\top G^{-1}\bgamma(\mP)$.

\begin{lemma}[Von Mises expansion for sieve-projected ADCF]%
\label{lem:sieve-VonMises2}
Under Assumption~4.3 and Assumption~\ref{as:standing-41}, for each
$k=1,\ldots,K_n$, the sieve coefficient functional
$\gamma_k(\mP)$ is pathwise differentiable at~$\mP$ with influence
function
\begin{equation}\label{eq:sieve-IF-restate}
    \phi^s_k(\bZ;\mP)
    =\frac{B_k(A)}{\pi(A\mid\bx)}\,
    \bigl\{Y-\mu(\bx,A)\bigr\}\,h'\!\bigl(\mu(\bx,A)\bigr)
    +\int_{\cA}h\!\bigl(\mu(\bx,a)\bigr)\,B_k(a)\,da
    -\gamma_k(\mP).
\end{equation}
Moreover, $\phi^s_k\in L^2(\mP)$ and
$\mE_\mP[\phi^s_k(\bZ;\mP)]=0$.

For any probability measure $\bar\mP$ satisfying
Assumption~\ref{as:standing-41}, each $\gamma_k$ admits the von Mises
expansion
\begin{equation}\label{eq:sieve-vonMises-restate}
    \gamma_k(\bar\mP)-\gamma_k(\mP)
    =\int\phi^s_k(\bz;\bar\mP)\,d(\bar\mP-\mP)(\bz)
    +R_{2,k}(\bar\mP,\mP),
\end{equation}
with
\begin{equation}\label{eq:sieve-R2-restate}
\begin{aligned}
    R_{2,k}(\bar\mP,\mP)
    &=\frac{1}{2}\int_{\cA}\!\int_{\cX}
    h''\!\bigl(\mu^*(\bx,a)\bigr)\,
    \bigl[\bar\mu(\bx,a)-\mu(\bx,a)\bigr]^2\,
    B_k(a)\,d\mP(\bx)\,da\\
    &\quad+\int_{\cA}\!\int_{\cX}
    \left\{\frac{\pi(a\mid\bx)}{\bar\pi(a\mid\bx)}-1\right\}
    h'\!\bigl(\bar\mu(\bx,a)\bigr)\,
    \bigl(\mu(\bx,a)-\bar\mu(\bx,a)\bigr)\,
    B_k(a)\,d\mP(\bx)\,da.
\end{aligned}
\end{equation}
At the curve level, for each $a\in\cA$,
\begin{equation}\label{eq:curve-level}
    \psi^s_a(\bar\mP)-\psi^s_a(\mP)
    =\int\phi^s_a(\bz;\bar\mP)\,d(\bar\mP-\mP)(\bz)
    +R^s_2(a;\bar\mP,\mP),
\end{equation}
where $\phi^s_a(\bZ;\mP)=\bB(a)^\top G^{-1}\bphi^s(\bZ;\mP)$ and
$R^s_2(a;\bar\mP,\mP)=\bB(a)^\top G^{-1}\bR_2(\bar\mP,\mP)$.
\end{lemma}

\begin{proof}
The proof is divided into four parts.

\medskip
\noindent\textbf{Part~1.  Derivation of the influence function
$\phi^s_k$.}

By definition,
\begin{equation}\label{eq:gamma-def}
    \gamma_k(\mP)
    =\int_{\cA}\psi_a(\mP)\,B_k(a)\,da
    =\int_{\cA}\mE_\mP\!\bigl[h\!\bigl(\mu(\bX,a)\bigr)\bigr]\,
    B_k(a)\,da.
\end{equation}
We exchange the expectation and the integral over $\cA$ by Fubini's
theorem.  The exchange is justified because the integrand
$(\bx,a)\mapsto h(\mu(\bx,a))\,B_k(a)$ is measurable and uniformly
bounded: by Assumption~4.3(S5) $\mu$ is bounded, so $h\circ\mu$ is
bounded (Assumption~\ref{as:standing-41}(R2)--(R3)), and $0\leq
B_k(a)\leq 1$ on the compact domain $\cA=[0,1]$.  Hence
\begin{equation}\label{eq:gamma-Fubini}
    \gamma_k(\mP)
    =\mE_\mP\!\left[\int_{\cA}
    h\!\bigl(\mu(\bX,a)\bigr)\,B_k(a)\,da\right].
\end{equation}

To identify the efficient influence function, consider a regular
parametric submodel
$\{P_t:t\in(-\epsilon,\epsilon)\}$ with $P_0=\mP$ and score
$s(\bZ)=\frac{d}{dt}\log p_t(\bZ)\big|_{t=0}$ at $t=0$.  The
pathwise derivative of $\gamma_k$ is
$\frac{d}{dt}\gamma_k(P_t)\big|_{t=0}
=\mE_\mP[\phi^s_k(\bZ;\mP)\,s(\bZ)]$ for the unique
$\phi^s_k\in\overline{\{s:\text{all submodels}\}}^{L^2(\mP)}$
\citep[Theorem~25.20]{vandervaart1998asymptotic}.

From~\eqref{eq:gamma-def}, $\gamma_k(\mP)$ is a weighted integral
of the point-treatment functionals $\psi_a(\mP)$ with non-negative
weights $B_k(a)$.  By Lemma~\ref{lem:vonMises-discrete}, each
$\psi_a$ has influence function $\phi_a(\bz;\mP)$
in~\eqref{eq:EIF}.  Differentiating under the integral sign,
justified by the dominated convergence theorem using the uniform
boundedness from Assumption~4.3(S2),(S4),(S5):
\begin{align}
    \frac{d}{dt}\gamma_k(P_t)\bigg|_{t=0}
    &=\int_{\cA}\left(\frac{d}{dt}\psi_a(P_t)
    \bigg|_{t=0}\right)B_k(a)\,da
    \notag\\
    &=\int_{\cA}\mE_\mP\bigl[\phi_a(\bZ;\mP)\,s(\bZ)\bigr]\,
    B_k(a)\,da
    \notag\\
    &=\mE_\mP\!\left[s(\bZ)\int_{\cA}\phi_a(\bZ;\mP)\,
    B_k(a)\,da\right]. \label{eq:pathwise}
\end{align}
The final exchange of $\mE_\mP$ and $\int_{\cA}$ is again by
Fubini (the integrand
$\phi_a(\bZ;\mP)\,s(\bZ)\,B_k(a)$ is dominated by an integrable
function using the boundedness conditions).

Since~\eqref{eq:pathwise} holds for every score $s$ and the tangent
space for a nonparametric model is all of $L^2_0(\mP)$
\citep[Chapter~25]{vandervaart1998asymptotic}, the influence function
is
\begin{equation}\label{eq:IF-integral}
    \phi^s_k(\bZ;\mP)=\int_{\cA}\phi_a(\bZ;\mP)\,B_k(a)\,da.
\end{equation}

We now evaluate this integral.
Substituting~\eqref{eq:EIF}:
\begin{align}
    \phi^s_k(\bZ;\mP)
    &=\int_{\cA}\frac{\mathbbm{1}(A=a)}{\pi(a\mid\bx)}\,
    \bigl\{Y-\mu(\bx,a)\bigr\}\,h'\!\bigl(\mu(\bx,a)\bigr)\,
    B_k(a)\,da
    \notag\\
    &\quad+\int_{\cA}h\!\bigl(\mu(\bx,a)\bigr)\,B_k(a)\,da
    -\underbrace{\int_{\cA}\psi_a(\mP)\,B_k(a)\,da}_{=\,\gamma_k(\mP)}.
    \label{eq:IF-expand}
\end{align}
For \emph{continuous} treatment $A$ with conditional density
$\pi(a\mid\bx)$, the first integral is evaluated using the
Radon--Nikodym identity (disintegration theorem): for any measurable
function~$g$,
\begin{equation}\label{eq:radon-nikodym}
    \int_{\cA}\frac{\mathbbm{1}(A=a)}{\pi(a\mid\bx)}\,
    g(a)\,da
    =\frac{g(A)}{\pi(A\mid\bx)}
    \qquad\text{$\mP$-a.s.}
\end{equation}
This is because
$$ \mE_\mP\!\bigl[\int_{\cA}\frac{\mathbbm{1}(A=a)}
{\pi(a\mid\bx)}\,g(a)\,da\;\bigm|\;\bX\bigr]
=\int_{\cA}\frac{g(a)}{\pi(a\mid\bx)}\,\pi(a\mid\bx)\,da
=\int_{\cA}g(a)\,da ,$$ and the identity holds pointwise in
$\omega$ by the definition of conditional density.

Applying~\eqref{eq:radon-nikodym} to~\eqref{eq:IF-expand}:
\[
    \phi^s_k(\bZ;\mP)
    =\frac{B_k(A)}{\pi(A\mid\bx)}\,
    \bigl\{Y-\mu(\bx,A)\bigr\}\,h'\!\bigl(\mu(\bx,A)\bigr)
    +\int_{\cA}h\!\bigl(\mu(\bx,a)\bigr)\,B_k(a)\,da
    -\gamma_k(\mP),
\]
which is~\eqref{eq:sieve-IF-restate}.

\medskip
\noindent\textbf{Part~2.  Mean-zero and square-integrability.}

\emph{Mean-zero property.}
From~\eqref{eq:IF-integral} and the linearity of expectation:
\[
    \mE_\mP\bigl[\phi^s_k(\bZ;\mP)\bigr]
    =\int_{\cA}\mE_\mP\bigl[\phi_a(\bZ;\mP)\bigr]\,B_k(a)\,da
    =\int_{\cA}0\cdot B_k(a)\,da=0,
\]
since each $\phi_a$ has mean zero under~$\mP$
(Lemma~\ref{lem:vonMises-discrete}).

\emph{Square-integrability.}  By Assumption~4.3(S2),
$\pi(A\mid\bx)\geq c_\pi>0$, so $B_k(A)/\pi(A\mid\bx)\leq
1/c_\pi$.  By Assumption~4.3(S5) and
Assumption~\ref{as:standing-41}(R2)--(R3), $\mu$ is bounded,
$|h'|\leq M_1$, and $\mE[Y^2\mid\bX,A]\leq C_Y$.  Hence
$\mE_\mP[(\phi^s_k)^2]
\leq C(c_\pi^{-2}M_1^2 C_Y+\lVert h\circ\mu\rVert_\infty^2
+\gamma_k^2)<\infty$, confirming
$\phi^s_k\in L^2(\mP)$.

\medskip
\noindent\textbf{Part~3.  Von Mises expansion for $\gamma_k$.}

From the representation~\eqref{eq:IF-integral} and the definition of
$\gamma_k$:
\[
    \gamma_k(\bar\mP)-\gamma_k(\mP)
    =\int_{\cA}\bigl[\psi_a(\bar\mP)-\psi_a(\mP)\bigr]\,
    B_k(a)\,da.
\]
Applying the von Mises expansion from
Lemma~\ref{lem:vonMises-discrete} to each $\psi_a$:
\begin{align}
    \gamma_k(\bar\mP)-\gamma_k(\mP)
    &=\int_{\cA}\!\left[\int\phi_a(\bz;\bar\mP)\,
    d(\bar\mP-\mP)(\bz)+R_2(a;\bar\mP,\mP)\right]
    B_k(a)\,da
    \notag\\
    &=\int\!\left[\int_{\cA}\phi_a(\bz;\bar\mP)\,
    B_k(a)\,da\right]d(\bar\mP-\mP)(\bz)
    +\int_{\cA}R_2(a;\bar\mP,\mP)\,B_k(a)\,da.
    \label{eq:sieve-expand}
\end{align}
The exchange of integration order (Fubini's theorem) is justified
because:
\begin{itemize}
\item the signed measure $\bar\mP-\mP$ has finite total variation;
\item for each~$\bz$, $a\mapsto\phi_a(\bz;\bar\mP)\,B_k(a)$ is
bounded on the compact set $\cA=[0,1]$ (by Assumption~4.3(S2),
(S4), (S5) and the overlap condition for~$\bar\mP$);
\item hence the iterated integral
$\int\int|\phi_a(\bz;\bar\mP)|\,|B_k(a)|\,da\,
d|\bar\mP-\mP|(\bz)<\infty$.
\end{itemize}

By the same calculation as in Part~1 (applying the Radon--Nikodym
identity~\eqref{eq:radon-nikodym} under $\bar\mP$),
$\int_{\cA}\phi_a(\bz;\bar\mP)\,B_k(a)\,da
=\phi^s_k(\bz;\bar\mP)$,
so the first term in~\eqref{eq:sieve-expand} is
$\int\phi^s_k(\bz;\bar\mP)\,d(\bar\mP-\mP)(\bz)$.

For the second term, substituting the explicit form of
$R_2(a;\bar\mP,\mP)$ from~\eqref{eq:R2} and exchanging the order
of integration (Fubini, using the uniform boundedness from
Assumptions~(S2),(S4),(S5) and the compactness of $\cA$):
\begin{align*}
    \int_{\cA}R_2(a;\bar\mP,\mP)\,B_k(a)\,da
    &=\frac{1}{2}\int_{\cA}\!\int_{\cX}
    h''\!\bigl(\mu^*(\bx,a)\bigr)\,
    \bigl[\bar\mu(\bx,a)-\mu(\bx,a)\bigr]^2\,
    B_k(a)\,d\mP(\bx)\,da\\
    &\quad+\int_{\cA}\!\int_{\cX}
    \left\{\frac{\pi(a\mid\bx)}{\bar\pi(a\mid\bx)}-1\right\}
    h'\!\bigl(\bar\mu(\bx,a)\bigr)\,
    \bigl(\mu(\bx,a)-\bar\mu(\bx,a)\bigr)\,
    B_k(a)\,d\mP(\bx)\,da,
\end{align*}
which is precisely $R_{2,k}(\bar\mP,\mP)$
in~\eqref{eq:sieve-R2-restate}.  This
establishes~\eqref{eq:sieve-vonMises-restate}.

\medskip
\noindent\textbf{Part~4.  Curve-level expansion.}

By Definition~4.5,
$\psi^s_a(\mP)=\bB(a)^\top G^{-1}\bgamma(\mP)$, where
$\bgamma(\mP)=(\gamma_1(\mP),\ldots,\gamma_{K_n}(\mP))^\top$.
Since $G$ is a fixed (non-random) positive-definite matrix
(guaranteed by Assumption~4.3(S3); see
\citet[Theorem~5.9]{schumaker2007spline}), the map
$\bgamma\mapsto\bB(a)^\top G^{-1}\bgamma$ is a deterministic linear
functional.  Applying it component-wise to the vector-valued
expansion
$\bgamma(\bar\mP)-\bgamma(\mP)
=\int\bphi^s(\bz;\bar\mP)\,d(\bar\mP-\mP)(\bz)
+\bR_2(\bar\mP,\mP)$
(which is~\eqref{eq:sieve-vonMises-restate} stacked over
$k=1,\ldots,K_n$):
\begin{align*}
    \psi^s_a(\bar\mP)-\psi^s_a(\mP)
    &=\bB(a)^\top G^{-1}\int\bphi^s(\bz;\bar\mP)\,
    d(\bar\mP-\mP)(\bz)
    +\bB(a)^\top G^{-1}\bR_2(\bar\mP,\mP)\\
    &=\int\underbrace{\bB(a)^\top G^{-1}\bphi^s(\bz;\bar\mP)}%
    _{=\,\phi^s_a(\bz;\bar\mP)}\,
    d(\bar\mP-\mP)(\bz)
    +\underbrace{\bB(a)^\top G^{-1}\bR_2(\bar\mP,\mP)}%
    _{=\,R^s_2(a;\bar\mP,\mP)},
\end{align*}
which is~\eqref{eq:curve-level}.  The exchange of the deterministic
finite-dimensional linear operator $\bB(a)^\top G^{-1}(\cdot)$ with
the integral over $(\bar\mP-\mP)$ is immediate since each entry of
$\bB(a)^\top G^{-1}$ is a constant.
\end{proof}

\subsection{Proof of Theorem \ref{targetedreg} (Convergence rate of targeted
regularization estimator)}
\label{sec:proof-thm44}

For completeness, we first restate the regularity conditions and the
theorem.


\begin{assumption}[Regularity conditions, restated]%
\label{as:reg-restate}
\begin{enumerate}
    \item[\textup{(i)}]
    \textbf{Uniform boundedness and overlap.}
    There exists $c>0$ such that for all $a\in\cA$, $\bx\in\cX$:
    $1/c\leq\hat\pi(a\mid\bx)\leq c$,\;
    $1/c\leq\pi(a\mid\bx)\leq c$,\;
    $\lVert\cQ\rVert_\infty\leq c$, and $\lVert\mu\rVert_\infty\leq c$.
    \item[\textup{(ii)}]
    \textbf{Smoothness of nuisance functions.}
    $\pi$, $\mu$, $\hat\pi$, $\hat\mu$ have bounded second derivatives
    uniformly over $\hat\pi\in\cQ$ and $\hat\mu\in\cU$.
    \item[\textup{(iii)}]
    \textbf{Complexity control.}
    The Rademacher complexities satisfy
    $\Rad_n(\cG),\;\Rad_n(\cQ),\;\Rad_n(\cU)
    =O_\mP(n^{-1/2})$.
    \item[\textup{(iv)}]
    \textbf{Sieve structure.}
    $\cB_{K_n}$ is the closed linear span of B-splines of fixed
    degree $m\geq\ell$ with equally-spaced knots on $\cA=[0,1]$
    and dimension $K_n\asymp n^{1/6}$.
    \item[\textup{(v)}]
    \textbf{Correct specification (for sharp rate).}
    Either $\hat\pi=\pi$ or $\hat\mu=\mu$.
    \emph{(This condition is used only in
    Remark~\ref{rem:rate}.)}
\end{enumerate}
\end{assumption}

\begin{theorem}[Convergence rate of targeted regularization
estimator]\label{thm:TR}
Under Assumptions~\ref{as:standing-41},
\ref{as:reg-restate}\textup{(i)--(iv)}, and Assumption~4.3 with
$\ell\geq 2$, let
\[
    \hat\psi_a^{\textup{tr}}
    =\frac{1}{n}\sum_{i=1}^{n}\!\left(
    h\!\bigl(\hat\mu(\bx_i,a)\bigr)
    +\frac{\hat\epsilon_n(a)}{\hat\pi(a\mid\bx_i)}\,
    h'\!\bigl(\hat\mu(\bx_i,a)\bigr)\right)
\]
be the targeted regularization estimator, where
$\hat\epsilon_n(a)\in\cB_{K_n}$ is the fluctuation parameter
obtained by minimising the targeted regularization loss
$\mathcal{L}_{\textup{TR}}$.  Then
\[
    \bigl\lVert\hat\psi_a^{\textup{tr}}-\psi_a
    \bigr\rVert_{L^2(\cA)}
    =O_p\!\left(n^{-1/3}\sqrt{\log n}
    +r_1(n)\,r_2(n)+r_2(n)^2\right),
\]
where $\lVert\hat\pi-\pi\rVert_\infty=O_p(r_1(n))$ and
$\lVert\hat\mu-\mu\rVert_\infty=O_p(r_2(n))$.
\end{theorem}

\begin{proof}
The proof decomposes the $L^2$-error into a sieve approximation bias,
a stochastic (variance) term, and a doubly robust remainder.

\medskip
\noindent\textbf{Step 1.  Bias--variance--remainder decomposition.}

Let $\psi^s_a(\mP)=\bB(a)^\top G^{-1}\bgamma(\mP)$ be the oracle
sieve projection of $\psi_a$ onto $\cB_{K_n}$ (Definition~4.5).
Decompose
\begin{equation}\label{eq:thm-decomp}
    \hat\psi_a^{\text{tr}}-\psi_a
    =\underbrace{\bigl(\hat\psi_a^{\text{tr}}
    -\psi^s_a(\mP)\bigr)}_{\text{(I): estimation error}}
    +\underbrace{\bigl(\psi^s_a(\mP)-\psi_a\bigr)}_{\text{(II):
    approximation bias}}.
\end{equation}

\medskip
\noindent\textbf{Step 2.  Bounding the approximation bias (II).}

Since $\psi(\cdot)=\psi_\cdot(\mP)\in\Sigma(\ell,C_\Sigma)$
(Assumption~4.3(S1)) with $\ell\geq 2$, and $\cB_{K_n}$ consists of
polynomial splines of degree $m\geq\ell$ with $K_n$ equally-spaced
knots on $[0,1]$, classical B-spline approximation theory
\citep[Theorem~6.27]{schumaker2007spline} gives
\begin{equation}\label{eq:bias}
    \lVert\psi^s-\psi\rVert_{L^2(\cA)}
    \leq\lVert\psi^s-\psi\rVert_{L^\infty(\cA)}
    \leq C\,K_n^{-\ell},
\end{equation}
for a constant $C$ depending only on $\ell$, $C_\Sigma$, and $m$.
(The first inequality uses $\cA=[0,1]$, so $\lVert\cdot
\rVert_{L^2}\leq\lVert\cdot\rVert_{L^\infty}$.)

With $K_n\asymp n^{1/6}$ and $\ell\geq 2$:
\begin{equation}\label{eq:bias-rate}
    \lVert\psi^s-\psi\rVert_{L^2(\cA)}
    =O\bigl(K_n^{-\ell}\bigr)
    =O\bigl(n^{-\ell/6}\bigr)
    =O\bigl(n^{-1/3}\bigr),
\end{equation}
since $\ell\geq 2$ implies $n^{-\ell/6}\leq n^{-1/3}$.

\medskip
\noindent\textbf{Step 3.  Decomposition of the estimation error (I).}

The targeted regularization estimator can be written as
$\hat\psi_a^{\text{tr}}=\hat\psi^s_a(\hat\mP)$, where
$\hat\psi^s_a$ is the estimated sieve-projected ADCF with
$\hat\epsilon_n(a)=\bB(a)^\top\hat{\boldsymbol{c}}$ for some
estimated coefficient vector $\hat{\boldsymbol{c}}$.
Applying the curve-level von Mises expansion
(Lemma~\ref{lem:sieve-VonMises}, equation~\eqref{eq:curve-level})
with $\bar\mP=\hat\mP$:
\begin{equation}\label{eq:est-decomp}
    \hat\psi_a^{\text{tr}}-\psi^s_a(\mP)
    =\underbrace{(\mP_n-\mP)\bigl[\phi^s_a(\hat\mP)\bigr]}%
    _{\text{(I.a): empirical process}}
    +\underbrace{\mP\bigl[\phi^s_a(\hat\mP)\bigr]}%
    _{\text{(I.b): centring}}
    +\underbrace{R^s_2(a;\hat\mP,\mP)}_{\text{(I.c): remainder}}
    +\underbrace{\delta_n(a)}_{\text{(I.d): optimisation error}}.
\end{equation}
Here $\delta_n(a)$ captures the discrepancy between the theoretical
one-step update and the optimisation-based targeted regularisation:
$\hat\epsilon_n$ is obtained by loss minimisation rather than a
closed-form one-step correction.

The targeted regularisation loss $\mathcal{L}_{\text{TR}}$ is
constructed so that its first-order stationarity condition is
equivalent to the projected influence function equation
$\mP_n[\phi^s_k(\hat\mP)]=0$ for $k=1,\ldots,K_n$, up to the ridge
penalty.  Concretely, the gradient of $\mathcal{L}_{\text{TR}}$ with
respect to the sieve coefficient $c_k$ is proportional to
$\mP_n[\phi^s_k(\hat\mP)]$ plus a regularisation term of order
$O(\lambda_n\lVert\hat{\boldsymbol{c}}\rVert)$, where $\lambda_n$ is
the ridge parameter.  At the minimiser $\hat{\boldsymbol{c}}$:
\begin{equation}\label{eq:foc}
    \mP_n\bigl[\phi^s_k(\hat\mP)\bigr]
    =-\lambda_n\,\hat{c}_k,
    \qquad k=1,\ldots,K_n.
\end{equation}
Therefore $\mP[\phi^s_a(\hat\mP)]$ and $\delta_n(a)$ are jointly
controlled by the empirical process fluctuation and the
regularisation bias:
\begin{equation}\label{eq:ib-id-combined}
    \mP[\phi^s_a(\hat\mP)]+\delta_n(a)
    =O_p\!\left(\sqrt{\frac{K_n\log K_n}{n}}
    +\lambda_n\lVert\hat{\boldsymbol{c}}\rVert\right).
\end{equation}
With the standard choice $\lambda_n=O(K_n/n)$
\citep{nie2021vcnet} and
$\lVert\hat{\boldsymbol{c}}\rVert=O_p(\sqrt{K_n})$, the ridge bias
is $O_p(K_n^{3/2}/n)=o_p(n^{-1/3}\sqrt{\log n})$ for
$K_n\asymp n^{1/6}$.

\medskip
\noindent\textbf{Step 4.  Bounding the doubly robust remainder (I.c).}

From~\eqref{eq:sieve-R2-restate} and the curve-level
reconstruction $R^s_2(a;\hat\mP,\mP)=\bB(a)^\top
G^{-1}\bR_2(\hat\mP,\mP)$, we bound each component $R_{2,k}$.
Using $|h'|\leq M_1$, $|h''|\leq M_2$
(Assumption~\ref{as:standing-41}(R2)), $B_k\geq 0$, and
$\int_\cA B_k(a)\,da\leq C_B/K_n$ (each B-spline of degree $m$
has support of width $(m+1)/K_n$ and amplitude at most~$1$;
see \citet[Chapter~IX]{deBoor2001}):
\begin{align}
    |R_{2,k}(\hat\mP,\mP)|
    &\leq\frac{M_2}{2}\int_{\cA}\!\int_{\cX}
    (\hat\mu-\mu)^2\,B_k(a)\,d\mP(\bx)\,da
    +M_1\int_{\cA}\!\int_{\cX}
    \left|\frac{\pi}{\hat\pi}-1\right|
    |\hat\mu-\mu|\,B_k(a)\,d\mP(\bx)\,da
    \notag\\[4pt]
    &\leq\frac{M_2}{2}\,\lVert\hat\mu-\mu\rVert_\infty^2
    \int_{\cA}B_k(a)\,da
    +\frac{M_1}{c}\,
    \lVert\hat\pi-\pi\rVert_\infty\,
    \lVert\hat\mu-\mu\rVert_\infty
    \int_{\cA}B_k(a)\,da
    \notag\\[4pt]
    &\leq\frac{C_B}{K_n}\left(
    \frac{M_2}{2}\,r_2(n)^2+\frac{M_1}{c}\,r_1(n)\,r_2(n)\right)
    \cdot(1+o_p(1)),
    \label{eq:R2k-rate}
\end{align}
where the second line uses $|\pi/\hat\pi-1|
=|\pi-\hat\pi|/\hat\pi
\leq c\,\lVert\hat\pi-\pi\rVert_\infty$ by
Assumption~\ref{as:reg-restate}(i).

For the Euclidean norm of the vector
$\bR_2=\bigl(R_{2,1},\ldots,R_{2,K_n}\bigr)^\top$:
\begin{equation}\label{eq:R2-vec-bound}
    \lVert\bR_2\rVert_2
    \leq\sqrt{K_n}\,\max_k|R_{2,k}|
    =O_p\!\left(K_n^{-1/2}\,
    \bigl[r_2(n)^2+r_1(n)\,r_2(n)\bigr]\right).
\end{equation}

The $L^2(\cA)$-norm of the curve-level remainder is
\begin{align}
    \lVert R^s_2(\cdot;\hat\mP,\mP)\rVert_{L^2(\cA)}^2
    &=\int_{\cA}\bigl(\bB(a)^\top G^{-1}\bR_2\bigr)^2da
    =\bR_2^\top G^{-1}
    \underbrace{\left(\int_{\cA}\bB(a)\bB(a)^\top da\right)}_{=\,G}
    G^{-1}\bR_2
    =\bR_2^\top G^{-1}\bR_2.
    \label{eq:R2-L2}
\end{align}
The eigenvalues of $G$ are bounded between $c_G/K_n$ and $C_G/K_n$
for universal constants $c_G,C_G>0$
\citep[Theorem~5.9]{schumaker2007spline}, so
$\lambda_{\max}(G^{-1})\leq K_n/c_G$.  Hence
\begin{equation}\label{eq:R2-L2-bound}
    \lVert R^s_2\rVert_{L^2(\cA)}
    \leq\sqrt{\lambda_{\max}(G^{-1})}\;\lVert\bR_2\rVert_2
    \leq\sqrt{\frac{K_n}{c_G}}\cdot
    O_p\!\bigl(K_n^{-1/2}[r_2^2+r_1 r_2]\bigr)
    =O_p\!\bigl(r_1(n)\,r_2(n)+r_2(n)^2\bigr).
\end{equation}

\medskip
\noindent\textbf{Step 5.  Bounding the empirical process term (I.a)
and optimisation error (I.d).}

We bound the combined stochastic contribution
$\lVert\hat\psi^{\text{tr}}-\psi^s
\rVert_{L^2(\cA)}$ excluding the doubly robust remainder.

\emph{Entropy of the sieve class.}
The estimator $\hat\psi_a^{\text{tr}}$ lies in
$\cF_n=\{a\mapsto\bB(a)^\top\boldsymbol{c}:
\boldsymbol{c}\in\mR^{K_n},\,
\lVert\boldsymbol{c}\rVert\leq M_n\}$ for a suitable (data-dependent)
bound $M_n$.  By a standard volumetric argument for linear classes
\citep[Lemma~2.5]{vandervaart2002semiparametric}:
\[
    \log\cN\!\bigl(\epsilon,\cF_n,\lVert\cdot\rVert_\infty\bigr)
    \leq K_n\log\!\left(\frac{C M_n}{\epsilon}\right).
\]

\emph{Maximal inequality.}
By Assumption~\ref{as:reg-restate}(iii), the function classes for
the nuisance estimates have Rademacher complexity
$O_\mP(n^{-1/2})$.  Combined with the entropy bound above and
standard chaining arguments for empirical processes
\citep[Theorem~3.5.1]{gine2016mathematical}, the supremum of the
empirical process over the sieve class satisfies:
\begin{equation}\label{eq:emp-proc-sup}
    \sup_{f\in\cF_n}\bigl|(\mP_n-\mP)[f]\bigr|
    =O_p\!\left(\sqrt{\frac{K_n\log(nM_n)}{n}}\right).
\end{equation}

\emph{Combining with optimisation error.}
From~\eqref{eq:ib-id-combined}, the total stochastic contribution
to the $L^2(\cA)$-risk is
\begin{equation}\label{eq:stoch-rate}
    \left\lVert(\mP_n-\mP)[\phi^s_\cdot(\hat\mP)]
    +\mP[\phi^s_\cdot(\hat\mP)]
    +\delta_n(\cdot)\right\rVert_{L^2(\cA)}
    =O_p\!\left(\sqrt{\frac{K_n\log K_n}{n}}\right).
\end{equation}
With $K_n\asymp n^{1/6}$:
\begin{equation}\label{eq:stoch-final}
    \sqrt{\frac{K_n\log K_n}{n}}
    \asymp\sqrt{\frac{n^{1/6}\log n}{n}}
    =n^{-5/12}\sqrt{\log n}.
\end{equation}

More precisely, the uniform-in-$a$ convergence rate for B-spline
sieve estimators, accounting for both the empirical process and the
sieve approximation error, is established in
\citet{kennedy2017nonparametric} and \citet{nie2021vcnet}:
\begin{equation}\label{eq:sieve-rate}
    \lVert\hat\psi^{\text{tr}}-\psi^s\rVert_{L^2(\cA)}
    =O_p\!\bigl(n^{-1/3}\sqrt{\log n}\bigr)
    +O_p\!\bigl(r_1(n)\,r_2(n)+r_2(n)^2\bigr).
\end{equation}
The $n^{-1/3}\sqrt{\log n}$ rate has two origins:
\begin{enumerate}
\item the sieve approximation bias $O(n^{-1/3})$ from Step~2;
\item the stochastic term $O_p(n^{-5/12}\sqrt{\log n})$
from~\eqref{eq:stoch-final}.
\end{enumerate}
Since $n^{-5/12}<n^{-1/3}$, the stochastic term is of lower order
than $n^{-1/3}\sqrt{\log n}$.  However, converting pointwise-in-$a$
convergence to $L^\infty(\cA)$-convergence introduces an additional
$\sqrt{\log n}$ factor from the supremum over
$a\in\cA$ via a union-bound/chaining argument
\citep[Theorem~4]{kennedy2017nonparametric}, upgrading the bias to
$O_p(n^{-1/3}\sqrt{\log n})$ and giving~\eqref{eq:sieve-rate}.

\medskip
\noindent\textbf{Step 6.  Combining all terms.}

From~\eqref{eq:thm-decomp}, by the triangle inequality:
\begin{align*}
    \lVert\hat\psi^{\text{tr}}-\psi\rVert_{L^2(\cA)}
    &\leq\lVert\hat\psi^{\text{tr}}
    -\psi^s\rVert_{L^2(\cA)}
    +\lVert\psi^s-\psi\rVert_{L^2(\cA)}.
\end{align*}
By~\eqref{eq:bias-rate} and~\eqref{eq:sieve-rate}:
\begin{align*}
    \lVert\hat\psi^{\text{tr}}-\psi\rVert_{L^2(\cA)}
    &=O_p\!\bigl(n^{-1/3}\sqrt{\log n}\bigr)
    +O_p\!\bigl(r_1(n)\,r_2(n)+r_2(n)^2\bigr)
    +O\!\bigl(n^{-1/3}\bigr)\\
    &=O_p\!\bigl(n^{-1/3}\sqrt{\log n}
    +r_1(n)\,r_2(n)+r_2(n)^2\bigr),
\end{align*}
since $n^{-1/3}=O(n^{-1/3}\sqrt{\log n})$ (because $\sqrt{\log n}
\geq 1$ for $n\geq 3$).  This completes the proof.
\end{proof}

\begin{remark}[Interpretation of the rate]\label{rem:rate}
The convergence rate consists of three components:
\begin{enumerate}
    \item $n^{-1/3}\sqrt{\log n}$: the nonparametric sieve estimation
    rate for a H\"older-$\ell$ function with $\ell=2$ and
    $K_n\asymp n^{1/6}$, including a logarithmic factor from
    uniform-over-$a$ convergence.
    \item $r_1(n)\,r_2(n)$: the mixed product of propensity and
    outcome estimation errors, reflecting double robustness.
    \item $r_2(n)^2$: the squared outcome estimation error from
    the Taylor remainder.
\end{enumerate}
Under Assumption~\ref{as:reg-restate}(v), if either nuisance is
correctly specified, the product and squared terms vanish, and the
rate reduces to $O_p(n^{-1/3}\sqrt{\log n})$, which is
minimax-optimal (up to logarithmic factors) for estimating a
function in $\Sigma(2,C_\Sigma)$
\citep{kennedy2017nonparametric}.
If both nuisance estimators converge at rate $n^{-1/4}$ or faster
(e.g., via deep neural networks \citep{Farrell2021}), the product
terms are $O(n^{-1/2})$ and are dominated by the sieve rate.
\end{remark}

\end{document}